\pdfoutput=1
\documentclass[11pt]{article}

\usepackage[]{EMNLP2023}

\usepackage{times}
\usepackage{latexsym}

\usepackage[T1]{fontenc}
\usepackage[utf8]{inputenc}

\usepackage{microtype}

\usepackage{inconsolata}

\usepackage{algorithm}
\usepackage[noend]{algpseudocode}
\usepackage{amsmath}
\usepackage{amssymb}
\usepackage{array}
\usepackage{arydshln}
\usepackage{bm}
\usepackage{booktabs}
\usepackage{color}
\usepackage{colortbl}
\usepackage{fancyvrb}
\usepackage[T1]{fontenc}
\usepackage{graphicx}
\usepackage{multirow}
\usepackage{pifont}
\usepackage{stfloats}
\usepackage{xcolor}

\definecolor{right}{RGB}{0,128,96}
\definecolor{wrong}{RGB}{192,0,32}
\newcommand{\Right}[1]{\textcolor{right}{#1}}
\newcommand{\Wrong}[1]{\textcolor{wrong}{#1}}
\newcommand{\cmark}{\Right{\ding{51}}}
\newcommand{\xmark}{\Wrong{\ding{55}}}
\DeclareMathOperator*{\argmax}{arg\,max}

\newcommand{\Ds}{\mathcal{D}}

\newcommand{\loss}{\mathcal{L}}
\newcommand{\ds}[1]{#1}
\newcommand{\model}[1]{#1}
\newcommand{\ckpt}[1]{\texttt{#1}}

\definecolor{purp}{HTML}{791f87}

\definecolor{difcolor}{RGB}{0,0,192}

\newcommand{\methodname}{\textsc{Vera}}

\title{\includegraphics[scale=0.03]{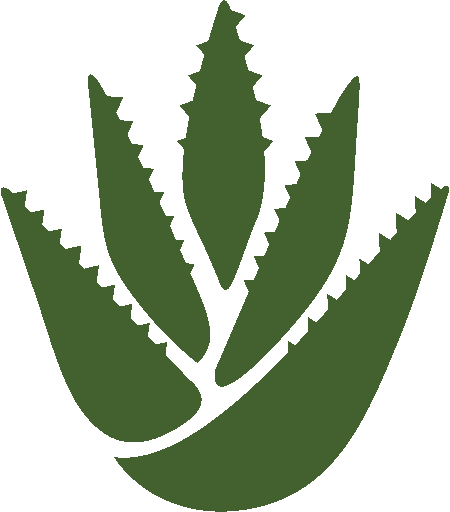} \methodname: A General-Purpose Plausibility Estimation Model for Commonsense Statements}

\newcommand{\aspace}{\hspace{0.4em}}
\newcommand{\uw}{$^{\heartsuit}$}
\newcommand{\aitwo}{$^{\spadesuit}$}
\newcommand{\harvard}{$^{\diamondsuit}$}
\newcommand{\ntu}{$^{\clubsuit}$}
\author{%
  Jiacheng Liu\uw$^*$ \aspace
  Wenya Wang\uw\ntu$^*$ \aspace
  Dianzhuo Wang\harvard \aspace \\
  \textbf{Noah A. Smith}\uw \aitwo \aspace
  \textbf{Yejin Choi}\uw \aitwo \aspace
  \textbf{Hannaneh Hajishirzi}\uw \aitwo \aspace \\
  \uw{}Paul G. Allen School of Computer Science \& Engineering, University of Washington \\
  \aitwo{}Allen Institute for Artificial Intelligence \aspace \ntu{}Nanyang Technological University \\
  \harvard{}John A. Paulson School of Engineering and Applied Sciences, Harvard University \\
  \texttt{liujc@cs.washington.edu} \aspace \small{$^*$Equal contribution.}
}

\begin{document}

\maketitle

\begin{abstract}
Today's language models can be remarkably intelligent yet still produce text that contains trivial commonsense errors. Therefore, we seek a retrospective verification approach that can reflect on the commonsense plausibility of the machine text, and introduce \methodname{}, a general-purpose model that learns to estimate the commonsense plausibility of declarative statements. % \alisa{personal preference but if you just deleted ``the first,'' it would still sound like the first one, and it's a little less blunt}
% \methodname is motivated by the observation that state-of-the-art LMs are prone to silly and unexpected failures in commonsense.
% \nascomment{this is a bit jarring; maybe continue with ``This verifier model is intended to address concerns about the reliability of text generated by language models, as their use grows.''}
% With the advent of remarkable capabilities of generative language models (LMs) and their growing use, concerns about the reliability and trustworthiness of their outputs have become more pressing.
% To date, even state-of-the-art LMs are prone to generating unreasonable \gary{or nonsensical?} claims, and such behavior is often left unguarded.
% We focus on the commonsense aspect of this problem,\alisa{this is redundant, you already say VERA is for commonsense} 
To support diverse commonsense domains, 
\methodname{} is trained on $\sim$7M commonsense statements that are automatically converted from 19 QA datasets and two commonsense knowledge bases, and using a combination of three training objectives. 
%, \methodname{} is a versatile model that effectively separates correct from incorrect statements across diverse commonsense domains.
% \nascomment{make this quantitative; if the methodology for measuring performance isn't established, point out that we introduce it}
When applied to solving commonsense problems in the verification format, \methodname{} substantially outperforms existing models that can be repurposed for commonsense verification, even including GPT-3.5/ChatGPT/GPT-4, and  %demonstrates strong %\nascomment{use numbers not value judgment words} \gary{there's not a single number to summarize everything, and so I'm worried it might cram too much detail into the abstract ... we have more concrete numbers in the intro} generalizability and calibration. \wenya{Consider changing to this: it further exhibits generalization capabilities to out-of-domain tasks and provides well-calibrated outputs.}
it further exhibits generalization capabilities to unseen tasks and provides well-calibrated outputs.
We find that \methodname{} excels at filtering machine-generated commonsense knowledge and is useful in detecting erroneous commonsense statements generated by models like \model{ChatGPT} in real-world settings.
\end{abstract}

\section{Introduction}
\label{sec:introduction}

We introduce \methodname{}, a general-purpose commonsense statement verification model.
This model is designed to estimate the plausibility of declarative, natural language statements based on commonsense knowledge.

\begin{figure}[t]
\centering
\includegraphics[width=0.9\linewidth]{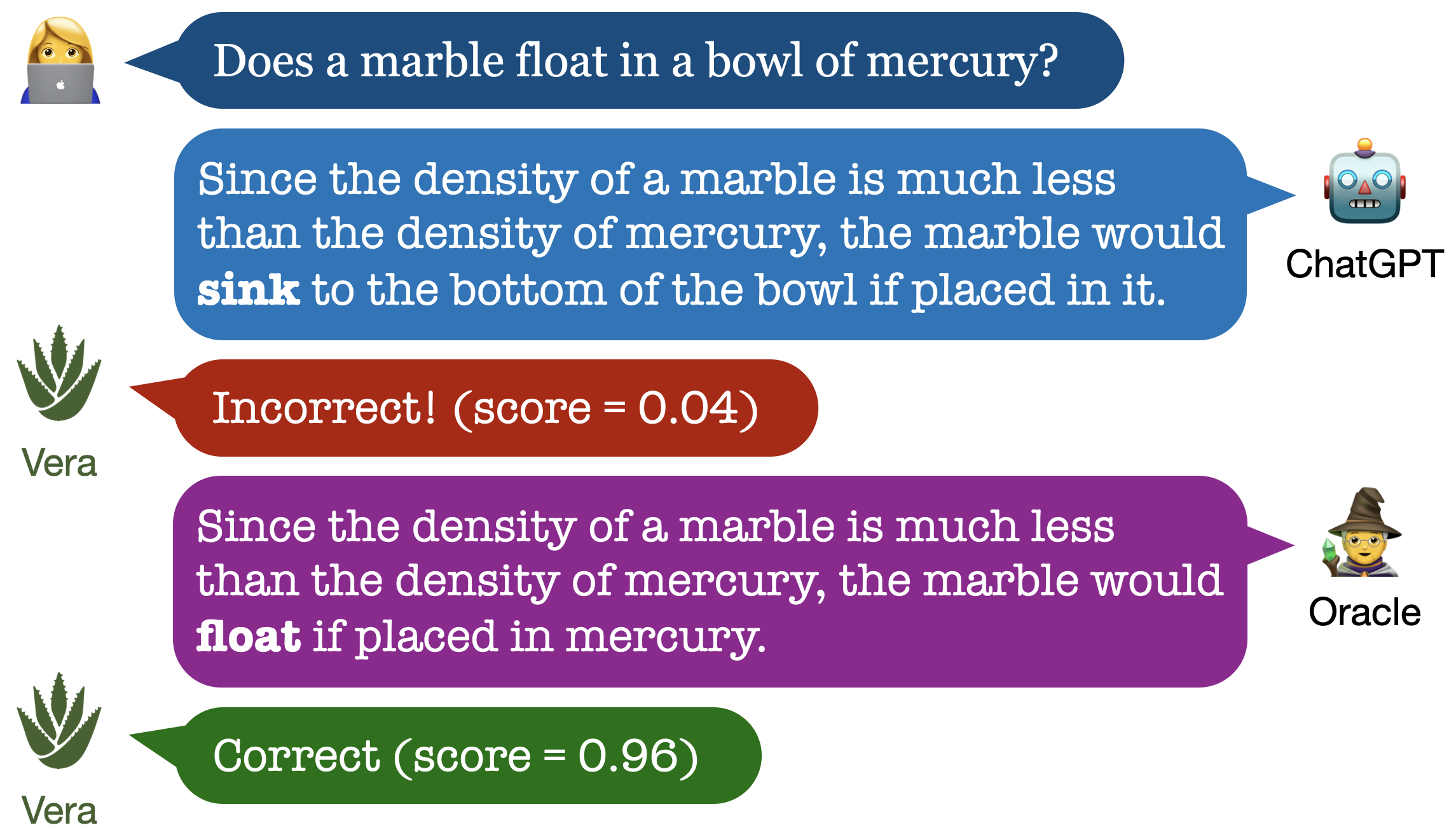}
\caption{
    \methodname{} estimates the correctness of declarative statements.
    Example adapted from a contribution made by Henry Minsky to \citet{1kDSERnROv5FgHbVN8z_bXH9gak2IXRtoqz0nwhrviCw} on February 23, 2023.
    % \alisa{I think it would be better if the hypothetical response didn't come from ChatGPT} \gary{We manually wrote this hypothetical response, it didn't come from ChatGPT.} \nascomment{so replace the ``ChatGPT'' label so people aren't misled}
}
\label{fig:teaser}
\vspace{-12pt}
\end{figure}

We build \methodname{} in response to the absence of good detectors of commonsense errors in text generated by language models (LMs).
% verification methods for texts generated by \nascomment{recommend ``language models'' throughout -- it's not relevant here that they are large} large language models (LLMs).
LMs have been advancing rapidly and have demonstrated remarkable success in various tasks, including question answering, natural language inference, sequence classification, and text generation.
% some can now pass professional exams, use tools, and coordinate other modules \nascomment{cite papers.  take care to stick to real results, don't just repeat hype}.
Yet these %extremely capable \nascomment{please do not say the models are intelligent, it contributes to hype and also contradicts the point you're trying to make!} \gary{how about seemingly intelligent} 
models still make simple commonsense mistakes.
As shown in \autoref{fig:teaser}, as of February 23, 2023, \model{ChatGPT} \citep{chatgptblog} reportedly output the text \textit{``since the density of a marble is much less than the density of mercury, the marble would sink to the bottom of the bowl if placed in it''}, which is clearly incorrect.
This kind of failure raises concerns about the reliability and trustworthiness of these models \citep{lin-etal-2022-truthfulqa}.%\alisa{can you cite existing work (e.g., Nelson's new paper) on this topic instead of giving a single example? sounds like bad scientific reasoning :/}
%A survey of open-source LMs shows that most \nascomment{I still think this should be cut.  first, what survey?  second, what is the link between left-to-right generation and "lacking an innate mechanism"?  I don't know what ``innate'' means here (or why it's important) or how we would test for such a mechanism.  also, this paper is not about left-to-right models vs. other models, so I don't see why we should bring this up.  mathematically, the left-to-right factorization does not restrict the set of models that can be represented by the architecture, because of the chain rule.}  generate text token-by-token from left to right, which implies that they lack an innate mechanism to reflect on their own outputs and make corrections when needed.

\methodname{} estimates a plausibility score for a commonsense statement %\nascomment{if we're evaluating the score rather than classification decisions, we should describe it as a scoring function instead} whether a commonsense statement is correct or incorrect 
based on its commonsense knowledge about the world.
It contrasts with \emph{fact} verification methods \citep{Thorne2018FEVERAL, Wadden2020FactOF}, which verify the correctness of claims based on evidence from a text corpus.
\methodname{} enables plausibility estimation where direct evidence is often not retrievable from some corpus, and usually some implicit, fuzzy reasoning is needed.
It operates solely with the commonsense knowledge stored in its model parameters, and does not have a retrieval component. %\nascomment{readers might take issue with our use of the term ``verification'' since that usually means either a proof of correctness (as in PL) or at least presenting evidence.} \wenya{I agree that ``verification'' has something to do with evidence, as in fact verification. How about replacing ``verification'' throughout the paper with ``scoring'' since we are now describing the model as a scorer.} \gary{Agree, let's change to ``estimating a plausibility score''.}

% \gary{Shall we move this to a separate, Problem Definition section?}
% \minor{
% \methodname{} 
% The statement is expressed in \textit{natural language}, not some symbolic format.
% The statement is \textit{declarative}, not a question.
% The statement should be \textit{self-contained}, 
% The statement should have an objective correctness value, like a \textit{proposition}, but according to the commonsense knowledge about our world, and the correctness value can be determined by commonsense.
% }

\methodname{} is built on top of \model{T5} \citep{Raffel2020t5}, a generic pretrained LM, by finetuning on a vast collection of correct and incorrect commonsense statements sourced from knowledge bases (KBs) and question answering (QA) datasets.
The 21 data sources (\autoref{tab:datasets}, appendix) amount to $\sim$7M statements encompassing a wide spectrum of domains, including general, scientific, physical, and social commonsense, as well as quantitative (reasoning about numbers) and qualitative (reasoning about qualitative relationships such as \textit{smaller}) commonsense.
% We employ multiple techniques to better train the verification model with such data.
We propose a novel two-stage training process %\nascomment{is this novel?  can't tell} 
that takes into account the scale and quality of data from different sources.
In addition to the standard multiple-choice binary classification objectives, we adopt a supervised contrastive loss \citep{Khosla2020SuperCL} to magnify the distinction between similar statements with different correctness labels. %\nascomment{not clear if this is novel} \alisa{I would make it clear what the novelty here is. are you the first to combine classification and contrastive loss?} \wenya{we are not the first to combine these two losses, so we use ``adopt'' here.}
%We also devise a method to use LMs to propose additional incorrect statements and thus augment the training data, and show that this improves model generalization. 
Furthermore, we propose an automatic way of augmenting the training data by eliciting LMs to generate incorrect answers to commonsense questions and empirically find it helps generalization.
%\nascomment{I like Wenya's version too, but should add that we find experimentally that it helps}
% , since LM-generated falsehoods can be more adversarial \nascomment{what does it mean that they can be more adversarial?} and they are shown to improve model generalizability \nascomment{citation?}.

% \methodname{} demonstrates outstanding \nascomment{subjective; either make a concrete comparative statement or give numbers} \gary{the concrete comparison depends on the application, I will add them in the bullet points below} performance in the following applications:
We evaluate \methodname{} in the following applications:
\begin{itemize}
\setlength{\itemsep}{0pt}
\setlength{\parskip}{0pt}
\setlength{\parsep}{0pt}
\item \textbf{Excelling in commonsense problems over GPT-series when repurposed for verification (\S\ref{sec:results_probsolving}).} \methodname{} can be applied to solve multiple-choice and boolean commonsense problems when expressed in the verification format, by scoring and ranking candidate hypotheses. It substantially outperforms existing models repurposed for commonsense verification (including \model{GPT-3.5}, \model{ChatGPT} and \model{GPT-4}), improving upon the best existing baseline, \model{Flan-T5}, with absolute improvement of 6\% on seen benchmarks and 4\% on unseen ones.
% \methodname{} also generalizes to some out-of-scope tasks, outperforming \model{Flan-T5} on multiple-choice story continuation, and is on par with \model{GPT-3.5} on entity-centric knowledge verification \wenya{check if this term is appropriate}.

\item \textbf{Filtering LM-generated commonsense knowledge (\S\ref{sec:results_knowledge}).} \methodname{} can filter noisy commonsense knowledge statements generated by other LMs, improving the effectiveness of LM-generated knowledge in downstream knowledge-augmented inferences. 
% outperforming existing critic models \citep{Bhagavatula2022I2D2IK}. 
% \alisa{deleted comparison to baseline bc I don't think it needs to be mentioned in this summary}
\methodname{} is well-calibrated, enabling filtering at customized thresholds. 
% Filtering with \methodname{} .

% \item \textbf{Solving out-of-domain problems (\S\ref{sec:results_ood}).} \alisa{this item name is too vague} \methodname{} has strong performance on tasks that are beyond the scope of its training data, such as
% % tasks that are slightly beyond its scope \alisa{then by definition, it's not beyond its scope. do you mean beyond the scope of the training data?}, such as 
% multiple-choice story continuation and simple fact verification, despite not being trained on those tasks. \alisa{how can vera generate stories? does ``verification'' also attach to ``story continuation''?} \wenya{changed it to multiple-choice story continuation.}

\item \textbf{Detecting commonsense errors in ChatGPT outputs (\S\ref{sec:results_chatgpt}).} Through a preliminary analysis, we find that \methodname{} can identify commonsense errors made by ChatGPT in-the-wild, with a precision of 91\% and a recall of 74\%.
An example of \methodname{} in action is shown in \autoref{fig:teaser}.
% in 9 out of 12 cases, \alisa{this seems like a tiny test bed, can you allude to how you collected these 12 cases and why they matter enough to base a section of your paper on it?} commonsense errors made by ChatGPT in-the-wild can be identified by \methodname{}.
\end{itemize}
% \nascomment{tie each of the above claims to sections in the paper}
% Aside from good performance, \methodname{} also exhibits good calibration, which can be useful in managing user's trust on its outputs. \nascomment{citations here}

% As we approach the multiple-choice QA problems under the verification setting, we also share some insights on QA-via-verification \nascomment{this makes me think you should have said something about ``verification as a task format'' earlier, perhaps it's one of your contributions?  this is kind of out of left field} compared with the standard QA format. \wenya{modified the above sentence to term it as ``QA-via-verification''.}
%Both have strengths and weaknesses, and can complement each other in creating better systems.\alisa{this last sentences almost doesn't say anything, I would just cut it} \nascomment{agree, unless you actually create better systems by combining them, in which case don't make the reader guess whether you did (``can'' is vague and will irritate readers)}
We hope \methodname{} can be a useful tool for improving the commonsense correctness of existing generative LM output and inspire more effort toward general-purpose and robust verification methods.

\section{Problem Definition and Scope}
\label{sec:task}

Our goal is to build a model that can estimate the plausibility of any given \textit{commonsense statement}.
The model takes as input a statement that (1) is expressed in \textbf{natural language}; (2) is \textbf{declarative}, as opposed to interrogative questions; (3) is \textbf{self-contained} without requiring additional context to comprehend; (4) has an objective, binary \textbf{correctness label}; and (5) in principle can be labeled using widely-held \textbf{commonsense knowledge} about the world. Encyclopedic knowledge (e.g., \textit{Ljubljana is
the capital of Slovenia.}) is out of scope.
Moving forward, unless explicitly noted, we use \textit{commonsense statement} to refer to statements within the above scope.
Though somewhat strict, this scope covers a broad range of potential applications.

For an input commonsense statement $x$, the model should output a real-valued score $s \in [0, 1]$
% \nascomment{this reinforces my view that we should frame it as a scoring function, not a classifier}
that represents its estimated plausibility of $x$.
While the gold correctness label is binary, we let the model output a score to reflect its confidence.
% This score can also be interpreted as the verifier's confidence in the correctness of $x$.
% Naively, the verifier outputs a binary label, \textit{correct} (1) or \textit{incorrect} (0). \nascomment{the way Wenya presented it, it seems better to just go straight to the score, and frame the whole paper around that.  so cut the preceding sentence?}
% Since we would like to know the verifier's confidence in its prediction, for an input commonsense statement $x$, we ask the verifier to output a real-valued score $s \in [0, 1]$ representing its confidence in the correctness of $x$.
A score of 1.0 means that it is completely confident that $x$ is correct, and a score of 0.0 means it is completely confident that $x$ is incorrect.
When predicting correctness label from the score, we use 0.5 as the threshold.

\section{Method}
\label{sec:method}

%\hanna{The section reads well, but you need to read the whole section sequentially to fully understand what's going on. We need an overview here explaining that a. you curate data that includes a collection of positive and related negative commonsense statements, b. you fine-tune an llm to this data, c. calibrate it. Then, in the data section, you should mention why commonsense statements do not naturally exist in the pretraining data, you use cs datasets and kb to find positive and importantly related negative statements, which enable trainign the model. Then in training you want to mention that you build batches based on postive and related negative examples (not clear how related negative examples mean; concretely define), then say you have a loss function that adds three loss, and explain them. }

In this section, we describe the whole pipeline to build \methodname{}. We start from curating large-scale training data including both correct and incorrect statements from diverse commonsense tasks (\S\ref{sec:data_construction}). Next, we learn a scoring model that takes a statement and returns a continuous score by finetuning a LM via 3 training objectives (\S\ref{sec:training}). An additional post hoc calibration strategy is applied to make the output scores well-calibrated (\S\ref{sec:calibration}).
%\hanna{please add a sentence explaining the pipeline; you take a large lm, fine-tune it in the data, and then refer to the corresponding sections. Here, we want the reader know what they are expecting to read in the next subsections e.g., what do you want to do with the data. }
% To build a general-purpose commonsense verifier, we propose an automatic pipeline of constructing a diverse collection of commonsense statement datasets, on which a pretrained language model could be finetuned.
% We further propose an effective training strategy that adds robustness for commonsense verification.

\subsection{Data Construction}
\label{sec:data_construction}

\begin{table}[t]
% \scriptsize
% \setlength{\tabcolsep}{3pt}
\centering
\resizebox{\linewidth}{!}{%
\begin{tabular}{c p{330pt} r}
\toprule
% \multicolumn{2}{c}{\textsc{Multiple-choice QA}} \\
% \midrule
\multirow{10}{*}{\rotatebox[origin=c]{90}{\textsc{Multiple-choice QA}}} & \textbf{Original example}: & \\
& \texttt{What would someone wear to protect themselves from a cannon?} & \\
& \texttt{(A) ungulate (B) bomber (C) body armor (D) tank (E) hat} & \\
& \texttt{Answer: (C)} & \\
\cmidrule{2-3}
& \textbf{Converted statement group}: & \\
& \Wrong{\texttt{One would wear an ungulate to protect themselves from a cannon.}} & \xmark \\
& \Wrong{\texttt{One would wear a bomber to protect themselves from a cannon.}} & \xmark \\
& \Right{\texttt{One would wear body armor to protect themselves from a cannon.}} & \cmark \\
& \Wrong{\texttt{One would wear a tank to protect themselves from a cannon.}} & \xmark \\
& \Wrong{\texttt{One would wear a hat to protect themselves from a cannon.}} & \xmark \\
\midrule
% \multicolumn{2}{c}{\textsc{Boolean QA}} \\
% \midrule
\multirow{5}{*}{\rotatebox[origin=c]{90}{\textsc{Boolean QA}}} & \textbf{Original example}: & \\
& \texttt{Can an average dog follow an instruction manual?} \\
& \texttt{Answer: no} \\
\cmidrule{2-3}
& \textbf{Converted statement group}: & \\
& \Wrong{\texttt{An average dog can follow an instruction manual.}} & \xmark \\
\midrule
% \multicolumn{2}{c}{\textsc{Knowledge base}} \\
% \midrule
\multirow{7}{*}{\rotatebox[origin=c]{90}{\textsc{Knowledge base}}} & \textbf{Original example}: & \\
& \texttt{Rubber stamps provide a way to make messages stand out.} & \\
\cmidrule{2-3}
& \textbf{Converted statement group}: & \\
& \Right{\texttt{Rubber stamps provide a way to make messages stand out.}} & \cmark \\
& \Wrong{\texttt{Arabic numbers provide a way to make messages stand out.}} & \xmark \\
& \Wrong{\texttt{Bandages provide a way to make messages stand out.}} & \xmark \\
& \Wrong{\texttt{Meat tenderizers provide a way to make messages stand out.}} & \xmark \\
\bottomrule
\end{tabular}
}%
\caption{
    Conversions from original commonsense QA problems and knowledge base entries to statement groups that are used for training.
}
\label{tab:conversion}
\vspace{-12pt}
\end{table}

% \hanna{motivate why you've taken this approach to collect data; it is worth mentioning that commonsense data are usually not explicitly mentioned in text in the wild; while using multiple choice commonsense datasets and KBs provide good resources to collect these statements. }
Labeled commonsense statements usually do not appear in text in the wild, while some commonsense question answering (QA) datasets and commonsense knowledge bases (KBs) are good sources for this kind of statements.
We collect correct and incorrect commonsense statements from the above two types of data source.
\autoref{tab:conversion} shows some examples on how these statements can be converted from QA problems and KB entries.
In total, we obtain $\sim$7M statements (for training) from 19 QA datasets (\S\ref{sec:data_qa}) and two KBs (\S\ref{sec:data_kb}) that encompass a wide spectrum of commonsense domains. % \hanna{ (section ... ) and commonsense knowledge bases (section...); also  remove the next part of the sentence; it has too much details.} including general, scientific, physical, and social commonsense, as well as quantitative commonsense which involves reasoning with numbers and qualitative commonsense which involves qualitative relationships such as \textit{smaller}. % \alisa{this is too repetitive with the introduction} \nascomment{I tightened a bit but don't know what the qualitative/quantitative things are or how they relate to the domains in the first list}
% \hanna{refer to table 2 early; maybe here} \gary{Done}
\autoref{tab:datasets} (appendix) lists these datasets with statistics.
All datasets we use are publicly available.
% Both contain large amounts of commonsense knowledge ranging from scientific facts to social interactions.
% Next, we detail our method on conversion and augmentation for each data source.

% \hanna{proposal: make these two paragraphs into 3.1.1 datasets 3.1.2 kbs, because the 2nd paragraph updates negative examples for the first paragraph, I believe.} \gary{done.}

\subsubsection{From Commonsense QA Datasets}
\label{sec:data_qa}

Numerous commonsense reasoning datasets have been published in recent years \citep{Davis2023BenchmarksFA}, and many of them are in the format of multiple-choice QA (selecting the correct answer out of a set of choices) or boolean (yes/no) QA.
These can be easily converted to correct and incorrect commonsense statements.
From multiple-choice QA problems, we combine the question and each answer choice to form declarative statements, which are correct when using the correct answer, and incorrect otherwise.
From boolean QA problems, we convert the question into a declarative statement, and keep the original label as the correctness label.
Concrete examples can be found in \autoref{tab:conversion}.

\paragraph{Statement groups.}
We refer to statements originating from the same problem as a \textit{statement group}.
Note that statement groups originating from multiple-choice problems contain at least two statements, of which one and only one is correct; statement groups originating from boolean problems contain only one statement, and it can be either correct or incorrect.

\noindent
We do conversion to declarative statements automatically. % ; the detailed process can be found in \S\ref{sec:conversion}.
% \alisa{imo this is all appendix material} \nascomment{agree, maybe just give a reference to that appendix section here}
From QA datasets, we create declarative statements from QA problems using the following method:
\begin{itemize}
\item If the problem contains a question, we convert the question and choice into a declarative statement using the question conversion model created by \citet{Chen2021CanNM}.
\item If the question is cloze-style, we replace the blank with the choice.
\item If the question is an incomplete sentence and the choice is a continuation to it, we concatenate the question and the choice.
\item If there is no question and the problem only asks to choose between some choices, we use the choice as the declarative statement.
\item For boolean problems, we always use \textit{yes} as the choice and create a single declarative statement for each problem. We use the original label as the correctness label of this statement.
\end{itemize}
In total, 19 commonsense QA datasets contribute $\sim$200k statement groups and $\sim$400k statements to the training set of \methodname{}.

% Answering questions about common sense has become a challenging task and has received growing attentions as a way of assessing language models' ability to understand and reason among their own knowledge of the real world beyond the given texts. For this purpose, several benchmark datasets in commonsense question answering have been introduced, most of which are formatted in a multiple-choice setting. An example of such format looks like this: \textit{Question: ``What happens when mercury is placed in water?'' (A) it dissolves; (B) it sinks; (C) it floats; (D) it hardens.} We select 19 datasets in total covering scientific, general, quantitative, social and physical domains. More details are shown in Table \ref{tab:datasets}. To adapt these questions in multiple-choice format for commonsense verification purpose, we convert each question with one of the choices to a declarative statement. For example, choice (A) for the above question becomes ``\textit{When mercury is placed in water, it dissolves.}'' Following \citet{Chen2021CanNM}, we perform this conversion via a fine-tuned LM, namely T5-3b trained on annotated input-outputs pairs \citep{demszky2018transforming} where the input consists of a question and an answer choice and the output corresponds to its declarative form. For each question, we treat the converted declarative sentence from the gold answer as a valid commonsense statement, whereas those converted from the incorrect choices are labeled as invalid commonsense statements.

\paragraph{LM-augmented falsehoods.}
Existing commonsense QA datasets are mostly manually constructed or assembled from standard school exams.
A model trained on these datasets might overfit specific annotation patterns from humans which may limit generalization.
Therefore, we augment QA problems with LM-generated answers and construct additional incorrect statements.
Specifically, for a multiple-choice question, we use a small LM to sample 50 possible answers to the question, and select the 3 least probable answers with generation probability less than 0.15 (making these unlikely to be correct answers). %\alisa{not sure I understand the point of the threshold}. \nascomment{some reviewers will wonder if you did analysis of a sample of these to see what fraction of them are actually false}
This threshold is chosen based on manual inspection over a small portion of examples.
We observe generated answers with probability larger than 0.15 are more likely to be plausible. %\wenya{added the above explanation.}
% We use greedy decoding to generate \nascomment{I don't think you need all these symbols.  instead of $K$ give the actual number.  then just say that you take the three least probable generations with probabilities less than $\lambda$ (but give a value not a Greek letter).  if you need these symbols later on, then keep them, but I think they make the paragraph harder to read than it needs to be.}  $K$ sequences and rank them in ascending order $G=\{g_1,...,g_K\}$ according to their generation probabilities $p_1\leq p_2 \leq ... \leq p_K$ where $p_k$ is the probability of generating $g_k$. Then we select 3 least probable generations $\{g_1,g_2,g_3\}$ provided that $g_3 < \lambda$ with $\lambda$ serving as the threshold to make sure all selected generations have low probabilities of being correct answers.
We create LM-augmented falsehoods for the training set of 9 commonsense QA datasets, as noted in \autoref{tab:datasets} (appendix).

\subsubsection{From Commonsense KBs}
\label{sec:data_kb}

Commonsense KBs (e.g., \ds{Atomic2020} in \citet{Hwang2020COMETATOMIC2O}, and \ds{GenericsKB} in \citet{Bhakthavatsalam2020GenericsKBAK}) contain a large number of correct commonsense statements.
To create incorrect statements, we automatically perturb KB entries by replacing the subject with three random subjects that appear in the KB.
\autoref{tab:conversion} shows how to convert an entry in GenericsKB to a statement group containing four statements, three of which are augmented via perturbations. The perturbed statements are relatively easy to identify and may contain false negatives. %(i.e., statements that are labeled \alisa{there's no labeling no? I would just remove this parentheses, it's obvious what you mean} incorrect are actually correct)
As noted in \S\ref{sec:training_stages}, we use these KB-constructed statements in a separate training stage that precedes training with QA-constructed statements.
In total, two commonsense KBs contribute $\sim$1.6M statement groups and $\sim$6M statements to the training set of \methodname{}.

% \gary{Commenting out since this is mentioned in Training Objectives.}
% Since the perturbed statements are relatively easy to identify and may contain false negatives, we use the KB-derived data for a round of finetuning before finetuning with data derived from commonsense QA datasets.
% % \wenya{This is a bit confusing. ``Meta-finetuning'' seems easier to be confused with ``meta learning''. More explanations could be better.} \gary{changed to finetuning.}
% We refer to the finetuning with KB-derived data as \textit{Stage A} training, and finetuning with QA-generated data as \textit{Stage B} training.

% \hanna{Merge model architecture and training into one subsection call it fine-tuning llm to cs verificationd data; then subsubsection 1 would be architecture, the next one batching, the next one training.} \gary{Done.}
\subsection{Model Training}
\label{sec:training}

\subsubsection{Model Architecture}
\label{sec:model_arch}
%\hanna{reading this section, makes it important to have an overview description at the beginning of section 3 explaining what this model architecture is supposed to do. }

Given a statement $x$, \methodname{} outputs a real-valued score $s \in [0, 1]$. % \alisa{vera outputs this regardless of whether x actually falls under the specified scope}
As we will use a transformer-based LM as the backbone of \methodname{}, we first extract the input representation $\mathbf{h}$ by selecting the last hidden state corresponding to the EOS input token.
We choose EOS because it is capable to encode the entire input in both bidirectional encoder models (e.g., \model{T5}'s encoder) and left-to-right decoder models (e.g., \model{LLaMA}).
Then a linear layer projects $\mathbf{h}$ to a scalar logit $z$, followed by a sigmoid function $\sigma(\cdot)$ that transforms the logit into a score $s$.
Formally,
\begin{align*}
\mathbf{h} = f_{\text{LM}}(x), z = f_{\text{linear}}(\mathbf{h}), s = \sigma(z).
\end{align*}
For brevity, we use $\mathbf{h}(x)$, $z(x)$ and $s(x)$ to refer to the representation, logit and score of an arbitrary input $x$.

% \minor{
% We use the T5 encoder and LLaMA as the backbone LM.
% T5 encoder is a bidirectional encoder model, and LLaMA is a left-to-right decoder model.
% The T5 tokenizer tokenizes input so that it ends with the EOS token \texttt{</s>} (token ID = 1).
% We manually configured the LLaMA tokenizer so that its output ends with the EOS token \texttt{</s>} (token ID = 2), and does not contain the BOS token \texttt{<s>} (token ID = 1).
% } 

% \noindent \textbf{Model}. Given a declarative statement $s$, {\methodname} aims to learn a single representation, denoted as 
% \begin{equation}
%     \mathbf{h} = f_{\mathrm{enc}}(s)
% \end{equation} 
% via an encoding function $f_{\mathrm{enc}}$. All the encoder models, e.g., BERT \cite{devlin2019-bert} and RoBERTa \cite{liu2019roberta}, could directly serve as the encoding function. However, one of the limitations is that these models have not been extensively pretrained over large-scale datasets and diverse tasks, compared to encoder-decoder models such as T5 \cite{Raffel2020t5}, making them less effective and knowledgeable about the world. Hence, we choose to use the encoder of the T5 model to generate a representation for each token in statement $s$ and select the one corresponding to the last token as the final representation $\mathbf{h}$ for $s$. We further apply a linear layer on top of $\mathbf{h}$ which transforms the feature vector to a scalar, indicating the validity score of statement $s$: $a = f_{\mathrm{linear}}(\mathbf{h})$.

\subsubsection{Batching}
\label{sec:batching}

The data we construct consists of statements belonging to different statement groups.
For reasons we will describe in \S\ref{sec:training_obj}, we put all statements belonging to the same statement group into the same batch. % \hanna{this is not super clear; this is clear for mc datasets, but not clear for how you deal with mc KBs or boolean datasets.  Either here or in the dataset curation you want to mention that your data consists of different statement groups, where the positive statement comes from a positive example in a commonsense dataset or a commensense KB and the negative examples are from the wrong choices, llm-generated or from the purturbation of commonsense KB.} \gary{Added a note. The content of statement groups was mentioned in \S\ref{sec:data_construction}.}
Each batch may contain multiple complete statement groups.
We denote by $B_G$ the number of statement groups and $B_S$ the number of statements in total within a single batch.
We denote the statement groups as $\{X_j\}_{j=1}^{B_G}$, and the statements as $\{x_i\}_{i=1}^{B_S}$.
$\{X_j\}_{j=1}^{B_G}$ is a partition of $\{x_i\}_{i=1}^{B_S}$.
$y_i \in \{0, 1\}$ is the correctness label of $x_i$.

\subsubsection{Training Objectives}
\label{sec:training_obj}

% \hanna{This sentence should also be moved at the beginning of the training section; we  like to see why you are introducing all these three loss functions} \gary{resolved.}
The model is trained with a linear combination of three losses: a binary classification loss, a multi-class loss, and a supervised contrastive loss, $\loss = \alpha \loss_{\text{bin}} + \beta \loss_{\text{mc}} + \gamma \loss_{\text{ctr}}$, which we describe below.
%\begin{align*}
%\loss &= \alpha \loss_{\text{bin}} + \beta \loss_{\text{mc}} + \gamma \loss_{\text{ctr}}.
%\end{align*}

\paragraph{Binary classification loss.}
Naively, commonsense statement verification can be viewed as a binary classification task.
Under this setting, the loss is
\begin{align*}
\loss_{\text{bin}} &= -y_i \log s(x_i) - (1 - y_i) \log (1 - s(x_i)).
\end{align*}

\paragraph{Multi-class loss.}
We expect the model to be robust against nuances in commonsense statements.
Ideally, the model should be able to recognize opposite correctness labels for a group of seemingly similar statements in surface forms, such as statements created from different choices of the same question, or perturbed from the same piece of knowledge in a KB.
To achieve this goal, we treat each statement group as a multi-class classification problem, maximizing the log-likelihood of the single correct statement in the statement group after passing the logits through a softmax. % additionally include a multiple-choice cross-entropy loss that maximizes the score of the single correct statement among distractors within the same group, \nascomment{I think you could replace the next two equations with text that says you treat each statement group as a multi-class classification problem, minimizing the log-likelihood of the correct statement in the group after passing the $z$-scores through a softmax. } \hanna{the above explanation proposed by Noah is really important even if you keep the equations.} \gary{Let's keep the equations for consistency with other losses, I added an explanation above.}
Formally,
\begin{align*}
\loss_{\text{mc}} &= -\log \frac{\exp z(x_{j*})}{\sum_{c=1}^{C_j} \exp z(x_{jc})},
\end{align*}
where $x_{j*}$ is the correct statement in $X_j$.
Note that the multi-class loss is not applicable to statement groups with only one statement (i.e., statement groups from boolean QA datasets). We empirically find that the multi-class loss indeed improves generalization towards unseen multiple-choice QA datasets as indicated in \autoref{fig:results_ablations} (appendix).

% \noindent \textbf{Objective}. Naively, commonsense statement verification can be viewed as a binary classification task where 1 indicates true statement and 0 indicates false statement. Under this setting, the objective becomes:
% \begin{equation}
%     l^b_i = -y_i \log \sigma(a_i) - (1-y_i)\log (1-\sigma(a_i)),
% \end{equation}
% where $y_i$ and $a_i$ correspond to the validity label and score for the $i$-th instance, respectively. Besides simple binary prediction, we expect the model to be more robust against nuances in commonsense statements. Ideally, the model should be able to recognize opposite validity labels for a group of seemingly similar statements in surface forms, such as statements converted from the same question or a KB fact with its augmented false statements. To achieve this goal, we additionally incorporate a multi-class cross entropy loss that tries to optimize the score of the single true statement among the other distractors within the same group. We denote it as $l^m$:
% \begin{equation}
%     l^m_j = - \log \frac{\exp (a_{j, c_j})}{\sum_{c=1}^{C} \exp (a_{j, c})}.
% \end{equation}
% Here $c_j$ refers to the index of the true statement in the $j$-th group. $a_{j, c_j}$ then corresponds to the score for the true statement in the $j$-th group. $C$ indicates the total number of statements (true and false) in each group.

\paragraph{Supervised contrastive loss.}
% \todo{Add a schematic figure for this loss.}
It has been shown \citep{Khosla2020SuperCL} that supervised contrastive learning helps to improve model robustness and generalization against input variations. In light of this,
%To further encourage the model to learn more discriminative representations and enhance its generalizability
% \nascomment{are these goals widely associated with supervised contrastive losses?  I'm not sure if the first part of this sentence is a widely accepted premise, or total speculation, or somewhere in between.} \wenya{revised this part}
we further adopt supervised contrastive learning on top of the input representations $\mathbf{h}$. We show in \autoref{fig:results_ablations} (appendix) that the contrastive loss indeed improve generalization to unseen datasets.
For each anchor statement $x_i$ in a batch, the contrastive loss aims to maximize the similarity between $x_i$ and each other statement $x_p$ that has the same correctness label as $x_i$ (i.e., positive example).
At the same time, we push apart $x_i$ and other statements $x_n$ that has opposite correctness label as $x_i$ (i.e., negative example).
The supervised contrastive loss is
\begin{align*}
& \loss_{\text{ctr}} = \\
&\quad -\log \frac{\sum_{k \in \mathcal{P}(i)} \exp [\text{cos}(\mathbf{h}(x_i), \mathbf{h}(x_k)) / \tau]}{\sum_{k \in \mathcal{P}(i) \cup \mathcal{N}(i)} \exp [\text{cos}(\mathbf{h}(x_i), \mathbf{h}(x_k)) / \tau]},
\end{align*}
where $\tau$ is a temperature hyperparameter, $\text{cos}(\cdot, \cdot)$ refers to cosine similarity, $\mathcal{P}(i) \subseteq [B_S]$ is the index set of statements that are positive examples for $x_i$, and $\mathcal{N}(i) \subseteq [B_S]$ is the index set of statements that are negative examples for $x_i$.
Formally,
\begin{align*}
\mathcal{P}(i) &= \big\{ k \mid 1 \le k \le B_S, y_k = y_i, k \ne i \big\}, \\
\mathcal{N}(i) &= \big\{ k \mid 1 \le k \le B_S, y_k \ne y_i \big\}.
\end{align*}
% \begin{equation}
%     l^c_k = - \frac{1}{|\mathcal{P}|}\sum_{p\in \mathcal{P}} \log \frac{\exp (sim(\mathbf{h}_k, \mathbf{h}_{k,p}) / \tau)}{\sum_{n'\in\mathcal{N}\cup \mathcal{P}}\exp (sim(\mathbf{h}_k, \mathbf{h}_{k,n'}) / \tau)}.
% \end{equation}
% Here $\mathcal{P}$ and $\mathcal{N}$ denote the index set of positive examples and negative examples, respectively, in the same batch. $\tau$ is a temperature hyperparameter and $sim$ refers to cosine similarity. We combine the three losses described above to train {\methodname}.

\subsubsection{Two-Stage Training}
\label{sec:training_stages}

Since data sourced from KBs are larger in scale but more noisy than data sourced from QA datasets, we take a two-stage training approach.
In training stage A, we start from a pre-trained LM and train with data sourced from KBs.
In training stage B, we start from the model obtained in stage A and train with data sourced from QA datasets.
During experiments we found that this setting is better than single-stage training with either data source or a mixture of the two.

% \hanna{I think this section can be better named as inference, right? or inference and calibration?} \gary{Fixed}
\subsection{Inference and Calibration}
\label{sec:calibration}

An ideal plausibility estimation model should be calibrated, that is, its confidence in its predictions should be approximately equal to the actual frequency of correctness.
During early experiments, we found that \methodname{} tends to be overconfident.
Therefore, we apply a post hoc calibration on \methodname{}'s output.
Following the temperature scaling method introduced in \citet{Guo2017OnCO}, during inference we divide the model-predicted logit by a temperature $T$ before computing the score, that is,
\begin{align*}
\mathbf{h} = f_{\text{LM}}(x), z = f_{\text{linear}}(\mathbf{h}), \tilde{z} = z / T, s = \sigma(\tilde{z}).
\end{align*}
Note that no temperature scaling is applied during model training.

With predictions on a validation set $\Ds = \{(x_i, y_i)\}_{i=1}^{\Ds}$, we estimate $T$ that gives the minimal expected calibration error (ECE) \citep{Naeini2015ObtainingWC} on this validation set.
\autoref{eqn:ece} in \S\ref{sec:def_metrics} shows how ECE is computed.
In practice, we use the combined development sets of the seen datasets (\S\ref{sec:eval_and_metrics}) to estimate $T$, and the optimal $T$ becomes a parameter of \methodname{}.
Note that temperature scaling does not change the relative ordering of prediction scores, and thus the other performance metrics (e.g., accuracy) are not affected (see detailed explanation in \S\ref{sec:calibration_more}).

\section{Experimental Setup}
% \alisa{this section title makes me think it's the evaluation section}
\label{sec:experiments}

In this section, we provide more details of model training, the evaluation protocol and metrics, and describe the baseline models we benchmark.

\subsection{Training Details}
\label{sec:training_details}

\paragraph{Datasets.}
For training stage A, we use the $\sim$1.6M statement groups ($\sim$6M statements) sourced from two commonsense KBs; for training stage B, we use the $\sim$200k statement groups ($\sim$400k statements) sourced from 19 commonsense QA datasets.
% gather correct commonsense statements from \ds{Atomic2020} \citep{Hwang2020COMETATOMIC2O} and \ds{GenericsKB} \citep{Bhakthavatsalam2020GenericsKBAK}, and perturb them to produce incorrect statements according to \S\ref{sec:data_construction}.
% For Stage B training, we gather data from 19 existing multiple-choice and boolean QA datasets in the commonsense reasoning domain.
% \autoref{tab:datasets} shows these datasets and their statistics.
% In total, we gathered $\sim$1.6M statement groups and $\sim$6M statements for Stage A training, and $\sim$200k statement groups and $\sim$400k statements for Stage B training.
For each training stage, we mix the training sets of all datasets together, without any re-weighting.

% \paragraph{Citation for datasets.}
% Atomic2020 \citep{Hwang2020COMETATOMIC2O},
% GenericsKB \citep{Bhakthavatsalam2020GenericsKBAK},
% OpenBookQA \citep{Mihaylov2018CanAS},
% ARC \citep{Clark2018ThinkYH},
% AI2Science \citep{Clark2018ThinkYH},
% CommonsenseQA \citep{Talmor2019CommonsenseQAAQ},
% QASC \citep{Khot2019QASCAD},
% PhysicalIQA \citep{Bisk2019PIQARA},
% SocialIQA \citep{Sap2019SocialIC},
% Winogrande \citep{Sakaguchi2019WINOGRANDEAA},
% Com2Sense \citep{Singh2021COM2SENSEAC},
% SciQ \citep{Welbl2017CrowdsourcingMC},
% QuaRel \citep{Tafjord2018QuaRelAD},
% QuaRTz \citep{Tafjord2019QuaRTzAO},
% CycIC\footnote{The CycIC dataset and leaderboard are available at \url{https://leaderboard.allenai.org/cycic}.},
% ComVE \citep{Wang2020SemEval2020T4},
% CommonsenseQA 2.0 \citep{Talmor2021CommonsenseQA2E},
% Symbolic Knowledge Distillation's annotated dataset \citep{West2021SymbolicKD},
% I2D2's annotated dataset \citep{Bhagavatula2022I2D2IK},
% Winograd Schema Challenge \citep{Levesque2011TheWS},
% COPA \citep{Gordon2011SemEval2012T7},
% NumerSense \citep{Lin2020BirdsHF},
% PROST \citep{ArocaOuellette2021PROSTPR},
% Spatial Commonsense \citep{Liu2022ThingsNW},
% Rainier's annotated dataset \citep{Liu2022RainierRK},
% SWAG \citep{Zellers2018SWAGAL},
% HellaSwag \citep{Zellers2019HellaSwagCA},
% CODAH \citep{chen2019codah},
% Story Cloze Test \citep{Mostafazadeh2016ACA},
% $\alpha$NLI \citep{Bhagavatula2019AbductiveCR},
% StrategyQA \citep{Geva2021DidAU},
% CREAK \citep{Onoe2021CREAKAD}.

\paragraph{Models.}
We use two types of pretrained LMs as the backbone of \methodname{}: (1) the encoder of \model{T5} \citep{Raffel2020t5}, which is a bidirectional encoder model; (2) LLaMA \citep{Touvron2023LLaMAOA}, which is a left-to-right decoder model. 
% For T5 encoder, we train two versions \methodname{}: one starting from \ckpt{Flan-T5-XXL}\footnote{\url{https://huggingface.co/google/flan-t5-xxl}} for best performance, and one starting from \ckpt{T5-v1.1}\footnote{\url{https://huggingface.co/google/t5-v1_1-xxl}} for less data contamination; both contain approximately 6B parameters.
For the \model{T5} encoder, we start from the pretrained \ckpt{T5-v1.1-XXL}\footnote{\url{https://huggingface.co/google/t5-v1_1-xxl}} whose encoder has about 5B parameters, and refer to the resulting model as \methodname{}-\model{T5}.
(During experiments we found that starting from \ckpt{Flan-T5-XXL}\footnote{\url{https://huggingface.co/google/flan-t5-xxl}} performs slightly worse than starting from \ckpt{T5-v1.1-XXL}.)
For \model{LLaMA}, we start from the pretrained \ckpt{LLaMA-7B} and refer to the resulting model as \methodname{}-\model{LLaMA}.
As we will see, \methodname{}-\model{T5} has better performance than \methodname{}-\model{LLaMA}, so unless explicitly specified, when we say \methodname{} we mean \methodname{}-\model{T5}.
See \autoref{tab:hypers} (appendix) for the complete hyperparameter settings and \S\ref{sec:app-exp-setup} for the implementation details.

\subsection{Evaluation and Baselines}
\label{sec:eval_and_metrics}

\paragraph{Evaluation protocol.}
We divide our evaluation into two parts:
(1) \textit{Seen} benchmarks, whose training set is used for model training.
(2) \textit{Unseen} benchmarks, whose training set is not used for model training.
We futher divide up the unseen benchmarks into \textit{type 1} and \textit{type 2}, where in type 1 benchmarks the task is similar to those in the seen benchmarks, while type 2 benchmarks are a bit further away in terms of the nature of the task.
Examples of type 2 unseen benchmarks include \ds{HellaSwag} which is contextualized with event descriptions, and \ds{CREAK} which involves reasoning among different entities.

\noindent
% During evaluation, we ask \methodname{} to assign a score to each commonsense statement of interest, and select those that receive the highest score(s).
Depending on the nature of the evaluation benchmark, we use different metrics to evaluate our model's performance.
Unless explicitly said otherwise, we report performance on the development set, where the gold labels are available, and we do not use the development sets of unseen datasets for model selection.
The overall metric reported over multiple benchmarks is the unweighted average of the metric over all these benchmarks, which accounts for the differently-sized evaluation sets.

% \nascomment{we don't need to give formulas for accuracies.  I suggest replacing the rest of this subsection with a short paragraph:}
\paragraph{Metrics.}
We report accuracy for multiple-choice and balanced boolean benchmarks.
For those unbalanced boolean benchmarks (e.g., LM-generated knowledge filtering datasets), we report area under the ROC curve (AUROC) and average precision (AP). % for true statements \nascomment{that's a bit fuzzy to me; is it a convention?}. 
To measure how well the model-predicted scores reflect confidence, we measure the ECE \citep{Naeini2015ObtainingWC} on the boolean benchmarks, following \autoref{eqn:ece}.

\paragraph{Baseline Models.}
We compare \methodname{} with the best publicly available models that can be directly used or repurposed for commonsense statement verification.
Roughly in increasing order of performance, these models are: \model{SKD Critic} \citep{West2021SymbolicKD}, \model{I2D2 Critic} \citep{Bhagavatula2022I2D2IK}, \model{UnifiedQA-v2} \citep{Khashabi2022UnifiedQAv2SG}, \model{Entailer} \citep{Tafjord2022EntailerAQ}, \model{GPT-3.5} \citep{gpt35docs}, \model{ChatGPT} \citep{chatgptblog}, \model{GPT-4} \citep{gpt4}, and \model{Flan-T5} \citep{Chung2022ScalingIL}.
See more details in \S\ref{sec:baselines_more}.
% These models are described below, roughly in increasing order of performance.

\section{Evaluation Results}
\label{sec:results}

% \hanna{here, explain the experiments. you should mention you show the effectiveness of VERA in three scenarios: commonsense benchmarks, evaluating generated commonsense knowledge, and qualitative evaluation of gpt generated knowledge.} \gary{Done.}

In this section, we evaluate the ability of \methodname{} to estimate the plausibility of commonsense statements and compare it with the baseline models.
We show the effectiveness of \methodname{} in three scenarios: solving commonsense problems, filtering LM-generated commonsense knowledge, and detecting commonsense errors in \model{ChatGPT} outputs.

\subsection{Solving Multiple-Choice and Boolean Commonsense Problems}
\label{sec:results_probsolving}

\begin{table}[t]
% \centering
% This image is generated by `python visualize_acc.py`
% \includegraphics[width=0.95\linewidth]{images/results_acc.png}
\setlength{\tabcolsep}{3pt}
\centering
\resizebox{\linewidth}{!}{%
\begin{tabular}{l ccc}   
\toprule
\textbf{Accuracy} & \textbf{Seen} & \textbf{Unseen (type 1)} & \textbf{Unseen (type 2)} \\
\midrule
SKD Critic (355M) & 36.64 & 38.34 & 43.40 \\
I2D2 Critic (355M) & 55.03 & 54.79 & 67.11 \\
UnifiedQA-v2 (11B) & 56.33 & 59.73 & 53.95 \\
Entailer (11B) & 73.79 & 71.47 & 70.72 \\
GPT-3.5 (175B) & 75.41 & 71.03 & 78.87 \\
ChatGPT$^\dagger$ & 62.11 & 61.20 & 62.83 \\
GPT-4$^\dagger$ & 72.35 & 77.40 & 70.29 \\
Flan-T5$^\ddagger$ (11B) & 79.50 & 77.62 & 78.89 \\
\midrule
\methodname{}-LLaMA (7B) & 82.99 & 75.51 & 82.56 \\
\methodname{}-T5 (5B) & \textbf{85.51} & \textbf{81.65} & \textbf{83.37} \\
\bottomrule
\end{tabular}
}%
% \vspace{8pt}
% This image is generated by `python visualize_boolean.py main`
% \includegraphics[width=0.95\linewidth]{images/results_boolean.png}
\caption{
    Results on problem-solving with \methodname{} on seen and unseen benchmarks.
    Average accuracy on the development sets is reported.
    Accuracy across different parts (seen, unseen (type 1), unseen (type 2)) are not directly comparable due to different underlying benchmarks.
    % For calibration curves, curves with saturated colors are results after applying post hoc
    % % \nascomment{I think you mean ``post hoc''?  :-)}
    % calibration (\S\ref{sec:calibration}), while curves with faded colors are results from the raw logits.
    % % \nascomment{no need to say this -- PR and ROC are more or less showing the same info:} Precision-recall curves are omitted here because the ordering of model performance follows similar trends as ROC curves.
    See \autoref{fig:results_probsolving_full} and \autoref{tab:results_seen}, \ref{tab:results_unseen}, \ref{tab:results_ood} (appendix) for full results.
    % \hanna{I think you should keep ROC curves and accuracies in 5.1 and 5.2, and then move the distribution and calibration curves to analyses. Since the patterns re similar between seen/unseen, you might keep one in the analyses and then move the rest to appendix.} \gary{Let's further compress the figures in the emnlp submission.}
    $\dagger$: The performance of \model{ChatGPT} and \model{GPT-4} may be under-estimated because we don't have access to the raw token logits.
    $\ddagger$: \model{Flan-T5} has been trained on some unseen benchmarks we use; see \autoref{tab:datasets_more} (appendix) for details on data contamination.
}
\label{fig:results_probsolving}
% \vspace{-12pt}
\end{table}

% \hanna{what is sota on these tasks? + do we report the results for individual datasets in appendix? } \gary{We can report the SOTA numbers if needed; results on individual datasets are in the appendix.}

The output plausibility scores from \methodname{} can be used for solving multiple-choice and boolean commonsense problems.
We first convert the problems into the statement group format (\S\ref{sec:data_construction}).
For multiple-choice problems, we choose the statement with the highest score in the statement group.
For boolean problems, we use $s = 0.5$ as the threshold to predict correctness labels of statements.

% \hanna{to clarify rewrite as: figure ... reports the results of VERA on multiple choice and boolean commonsense benchmarks. [probably you wan to add a sentence how VERA that returns plausiblity can be used to solve MC } \gary{Done.}
\autoref{fig:results_probsolving} reports the results when \methodname{} is applied to solve commonsense problems. See \autoref{fig:results_probsolving_full} and \autoref{tab:results_seen}, \ref{tab:results_unseen}, \ref{tab:results_ood} (appendix) for full results including AUROC and AP.  % \nascomment{the point isn't to solve the problems, though, is it?  it's to score for plausibility} \gary{addressed by adding the above paragraph.}
On seen benchmarks (16 multiple-choice and one boolean), \methodname{} outperforms the best baseline, \model{Flan-T5}, by 6\% on (absolute) accuracy and 9\% on AUROC.
\methodname{} beats \model{Flan-T5} by 4\% accuracy and 5\% AUROC on type 1 unseen benchmarks (four multiple-choice and one boolean), and by 4\% accuracy and 6\% AUROC on type 2 unseen benchmarks (five multiple-choice and two boolean), demonstrating good generalization.
% \hanna{figure 2 has reported too many results, but you haven't discussed them here.  I believe accuracy and ROC curves are results of mc choice or boolean, averaged over all datasets. you need to mention that these are averaged. The distribution of predictions is more like an ablation and analyses, right? you have not discusesd it. Also calibration curve is not discussed, I think that's also more like an ablation; you probaby can have two figures, one for ROC and accuracy to report the results, explained in one paragraph; 2nd paragraph explain analyses in distribution prediction and calibration curve.} \gary{The averaging is mentioned in \S\ref{sec:eval_and_metrics}. Added discussion of logit distribution and calibration.}
% It has good out-of-distribution generalizability, outperforming all baselines on unseen benchmarks in the multiple-choice format, and is on-par with the best baseline, \model{Flan-T5}, on unseen benchmarks in the boolean format (which constitutes only a single dataset, \ds{SpatialCS}).
% \wenya{Maybe we can remove either ROC curve or P-R curve in the main paper and put it to appendix? ROC and P-R seem to follow similar patterns. A small design issue: Is it possible to reposition the legend such that it won't interfer too much with the curve?} \gary{If we need to choose, I prefer to keep ROC curves here and move P-R curve into Appendix. The legends are a bit tricky, do you think it's okay if we omit the AUC numbers?} \wenya{Agreed.}
\methodname{}-\model{T5} has better performance than \methodname{}-\model{LLaMA} across the board, which may be due to its bidirectional connectivity.
Aside from performance, \methodname{} also has good calibration, with ECE no higher than 3\% on seen and unseen benchmarks.
The post hoc calibration method improves calibration across all three parts.

% \hanna{this shouldn't be a section; move this as a paragraph in section 5.1} \gary{Done}
% \nascomment{same issue as earlier with ``OOD''} \gary{changed to out-of-scope}
% \methodname{} also demonstrates good out-of-scope generalization.
% It outperforms all baselines on story-continuatation benchmarks (which coincides with the set of multiple-choice benchmarks in OOD), and is on-par with the strongest baselines on simple fact verification benchmarks (which coincides with the set of boolean benchmarks in OOD), despite never being trained to solve these tasks.

% \nascomment{this feels like a footnote in the results section before you introduce the accuracy results, not a subsection} \hanna{+1; just a footnote in your accuracy results; or in experimental details.}
Typically we may need to choose a threshold for binary classification in boolean datasets.
However, we notice that a zero logit ($z = 0$) is generally close to the optimal decision threshold between correct and incorrect commonsense statements.
Therefore we do not estimate a model-specific threshold, and simply use the default threshold: $z = 0$, or equivalently, $s = 0.5$.

\subsection{Filtering LM-generated Commonsense Knowledge}
\label{sec:results_knowledge}

\begin{figure}[t]
\centering
% This image is generated by `python visualize_knowledge.py main`
\includegraphics[width=\linewidth]{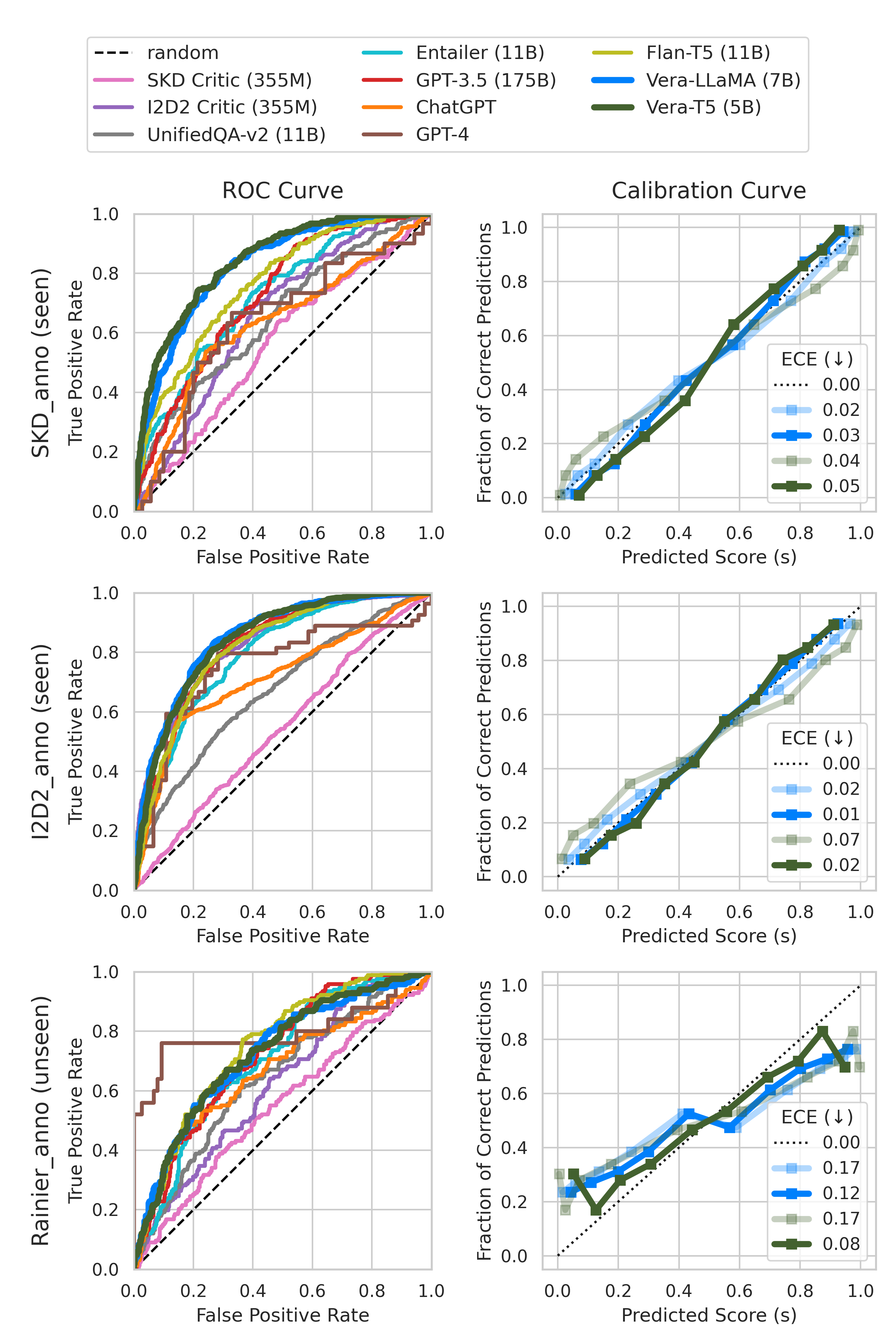}
\caption{
    Results for filtering LM-generated commonsense knowledge with \methodname{}.
    We plot the calibration curve for both the uncalibrated version (w/ faded color) and calibrated version (w/ saturated color) of the \methodname{} model.
    Results on the development sets are reported.
    % \nascomment{remove: Precision-recall curves are omitted here because the ordering of model performance follows similar trends as ROC curves.}
    See \autoref{fig:results_knowledge_full} for full results.
}
\label{fig:results_knowledge}
% \vspace{-12pt}
\end{figure}

\autoref{fig:results_knowledge} reports the results when \methodname{} is applied to filter LM-generated commonsense knowledge.
On the two seen benchmarks, \ds{SKD\_anno} and \ds{I2D2\_anno}, \methodname{} is a better knowledge filter than all baseline models, in terms of both AUROC and AP.
In particular, on \ds{I2D2\_anno} it outperforms the \model{I2D2 critic} model by 2\% AUROC, which is specifically trained on the \ds{I2D2\_anno} dataset and does not generalize well to other benchmarks.
On the unseen benchmark, \ds{Rainier\_anno}, \methodname{} is also comparable with the best baselines like \model{Flan-T5} and \model{GPT-3.5}.
As for calibration, the ECE is no higher than 8\% on all three benchmarks.

\begin{table}[t]
\setlength{\tabcolsep}{3pt}
\centering
% \begin{minipage}{0.49 \textwidth}
\resizebox{\linewidth}{!}{%
\begin{tabular}{lll ccc}
\toprule
\textbf{Generator} & \textbf{Filter} & \textbf{QA Model} & \textbf{Acc} & \textbf{Usefulness} & $\Delta$ \\
\midrule
-- & -- & UnifiedQA & 60.45 & -- & -- \\
GPT-3 & -- & UnifiedQA & 67.44 & +6.99 & -- \\
GPT-3 & \methodname{} & UnifiedQA & \textbf{70.67} & \textbf{+10.22} & \textbf{+46\%} \\
\midrule
-- & -- & UnifiedQA & 60.45 & -- & -- \\
Rainier & -- & UnifiedQA & 61.78 & +1.33 & -- \\
Rainier & \methodname{} & UnifiedQA & \textbf{64.88} & \textbf{+4.43} & \textbf{+233\%} \\
\bottomrule
\end{tabular}
}%
% \end{minipage}
% \hfill
% \begin{minipage}{0.49 \textwidth}
% \resizebox{\linewidth}{!}{%
% \begin{tabular}{lll ccc}
% \toprule
% \textbf{Generator} & \textbf{Filter} & \textbf{QA Model} & \textbf{Acc} & \textbf{Usefulness} & $\Delta$ \\
% \midrule
% -- & -- & UnifiedQA-large & 60.45 & -- & -- \\
% Rainier-large & -- & UnifiedQA-large & 61.78 & +1.33 & -- \\
% Rainier-large & \methodname{} & UnifiedQA-large & \textbf{64.88} & \textbf{+4.43} & \textbf{+233\%} \\
% \bottomrule
% \end{tabular}
% }%
% \end{minipage}
% \vspace{-4pt}
\caption{
    Results of introducing \methodname{} into the Generated Knowledge Prompting pipeline \citep{Liu2021GeneratedKP}.
    The QA model is \ckpt{UnifiedQA-large}, and the generator is either \ckpt{GPT-3 (davinci)} or \ckpt{Rainier-large} when applicable.
    Average accuracy on the development set is reported; see \autoref{tab:results_gpi_more} (appendix) for detailed results.
}
\label{tab:results_gpi}
% \vspace{-12pt}
\end{table}

% \begin{figure*}[!h]
% \centering
% \includegraphics[width=0.8\linewidth]{images/results_gpi_gpt3.png}
% \includegraphics[width=0.8\linewidth]{images/results_gpi_rainier.png}
% \caption{
%     Results of introducing knowledge filtering with \methodname{} into the Generated Knowledge Prompting pipeline \citep{Liu2021GeneratedKP}.
%     The QA model is \ckpt{UnifiedQA-large}.
%     Average accuracy on the development set is reported; see \autoref{tab:results_gpi_more} for detailed results.
% }
% \label{fig:results_gpi}
% % \vspace{-16pt}
% \end{figure*}

We find that filtering commonsense knowledge using \methodname{} can greatly improve the performance of knowledge-augmented reasoning methods.
In the Generated Knowledge Prompting framework \citep{Liu2021GeneratedKP}, when solving a commonsense QA problem, first a knowledge model generates several commonsense knowledge statements relevant to the question, and then a QA model makes predictions based on them.
A big problem that hinders the effectiveness of this framework is that model-generated knowledge is not always factual, and incorrect knowledge statements can mislead the QA model.
We introduce \methodname{} to filter these statements before passing them to the QA model.
In particular, we keep those statements that receive a score higher than 0.5 from \methodname{}.

Following \citet{Liu2022RainierRK}, we use \ckpt{UnifiedQA-large} as the QA model, and consider two knowledge models: few-shot \ckpt{GPT-3 (davinci)} \citep{Brown2020LanguageMA} and \ckpt{Rainier-large} \citep{Liu2022RainierRK}.
We follow the evaluation settings as in \citet{Liu2022RainierRK}, and for few-shot \ckpt{GPT-3 (davinci)}, we use the same task-specific few-shot prompts and same process to generate silver knowledge as in \citet{Liu2022RainierRK}.
% \nascomment{I think this would work better as a small table since it's only 6 numbers} \gary{Changed to table.}
Results are shown in \autoref{tab:results_gpi}.
% \nascomment{the way you've written it, it sounds like the numbers that follow are a summary of the figure.  but I don't see how these huge gains are reflected in the figure.  where is this coming from?} \gary{Added improvement numbers in the table.}
Applying knowledge filtering with \methodname{} increases the usefulness of \model{GPT-3}'s and \model{Rainier}'s knowledge by 46\% and 233\%, respectively.
\methodname{} can effectively supervise and improve the quality of commonsense knowledge generated by a much larger model, \ckpt{GPT-3 (davinci)}.
Detailed results (\autoref{tab:results_gpi_more}, appendix) show that there is increased effectiveness in every individual benchmark.

\subsection{Preliminary Study on Detecting Commonsense Errors made by ChatGPT} 
\label{sec:results_chatgpt}
% \hanna{this only includes a list of 19 examples; edit the title to something like qualitative evaluation on ChatGPT commonsesne errors} \gary{Expanded to 27 examples and included some quantitative numbers.}

% \input{floats/results_chatgpt_failures_metrics}
\begin{table*}[t]
\setlength{\tabcolsep}{3pt}
\centering
\resizebox{\textwidth}{!}{%
\begin{tabular}{l p{500pt} c c}
\toprule
\textbf{Date} & \textbf{\Wrong{Original} / \Right{Corrected}} & \textbf{Score} & \textbf{Pred} \\
\midrule
\multirow{2}{*}{2023/01/05} & \Wrong{It is possible for a solar eclipse to be followed by a lunar eclipse the next day.} & 0.86 & \cmark \\
& \Right{It is impossible for a solar eclipse to be followed by a lunar eclipse the next day.} & 0.48 & \xmark \\
% \midrule
% \multirow{2}{*}{2023/01/05} & In the statement "The trophy didn't fit in the suitcase because it was too small," the trophy is the object that is too small to fit in the suitcase. & 0.04 & \xmark \\
% & In the statement "The trophy didn't fit in the suitcase because it was too small," the suitcase is the object that is too small to fit the trophy in. & 0.98 & \cmark \\
\midrule
\multirow{2}{*}{2023/01/06} & \Wrong{The time it takes for a given number of cars to travel a fixed distance is directly proportional to the number of cars.} & 0.26 & \xmark \\
& \Right{The time it takes for a given number of cars to travel a fixed distance is invariant of the number of cars.} & 0.52 & \cmark \\
\midrule
\multirow{2}{*}{2023/01/06} & \Wrong{If A sits next to B and B sits next to C, then A must sit next to C.} & 0.20 & \xmark \\
& \Right{If A sits next to B and B sits next to C, then A may not sit next to C.} & 0.60 & \cmark \\
\midrule
\multirow{2}{*}{2023/01/10} & \Wrong{If two cats can eat two cans of food in a minute, then it would take six cats to eat three cans of food in a minute.} & 0.05 & \xmark \\
& \Right{If two cats can eat two cans of food in a minute, then it would take three cats to eat three cans of food in a minute.} & 0.67 & \cmark \\
% \midrule
% \multirow{2}{*}{2023/01/10} & The United States has had two black presidents: Barack Obama, who served two terms from 2009 to 2017, and Donald Trump, who served one term from 2017 to 2021. & 0.16 & \xmark \\
% & The United States has had one black president: Barack Obama, who served two terms from 2009 to 2017. & 0.91 & \cmark \\
% \midrule
% There exists a place on Earth where all directions are north. & 0.69 \\
% There exists a place on Earth where all directions are north. & 0.39 \\
\midrule
\multirow{2}{*}{2023/01/11} & \Wrong{A three-dimensional cube has eight faces.} & 0.46 & \xmark \\
& \Right{A three-dimensional cube has six faces.} & 0.70 & \cmark \\
\midrule
\multirow{2}{*}{2023/01/30} & \Wrong{It is possible to draw a diagonal line in a triangle.} & 0.80 & \cmark \\
& \Right{It is impossible to draw a diagonal line in a triangle.} & 0.28 & \xmark \\
\midrule
\multirow{2}{*}{2023/02/21} & \Wrong{70 is a smaller number than 58.} & 0.14 & \xmark \\
& \Right{70 is a larger number than 58.} & 0.85 & \cmark \\
\midrule
\multirow{2}{*}{2023/02/23} & \Wrong{Since the density of a marble is much less than the density of mercury, the marble would sink to the bottom of the bowl if placed in it.} & 0.04 & \xmark \\
& \Right{Since the density of a marble is much less than the density of mercury, the marble would float if placed in mercury.} & 0.96 & \cmark \\
\midrule
\multirow{2}{*}{2023/02/25} & \Wrong{Both a house and a pound of feathers weigh the same, which is one pound.} & 0.25 & \xmark \\
& \Right{A house weighs more than one pound, while a pound of feathers weighs one pound.} & 0.87 & \cmark \\
% \midrule
% -- & It is not possible for a person to marry their mother's daughter-in-law. & 0.25 & \xmark \\
% & It is possible for a person to marry his mother's daughter-in-law. & 0.53 & \xmark \\
% \midrule
% \multirow{2}{*}{--} & CPU computation is generally faster than GPU computation for deep learning. & 0.17 & \xmark \\
% & CPU computation is generally slower than GPU computation for deep learning. & 0.98 & \cmark \\
\bottomrule
\end{tabular}
}%
\caption{
    % \nascomment{not clear if these are randomly sampled or cherrypicked} \hanna{+1; cherrypicked? why only 19?} \gary{These are not cherry-picked, the distribution in these examples roughly equals the quantitative results.}
    Examples of commonsense mistakes made by ChatGPT, and how \methodname{} can detect them.
    In each section, the first line is the original, incorrect commonsense statement in ChatGPT's output, and the second line is the authors' manually corrected version of the statement. % \nascomment{corrected by who?}
    Each statement is followed by \methodname{}'s score and predicted correctness label.
    Examples are adapted from \citet{giuven95/chatgpt-failures, 1kDSERnROv5FgHbVN8z_bXH9gak2IXRtoqz0nwhrviCw, Borji2023ACA}.
}
\label{tab:results_chatgpt_failures}
% \vspace{-12pt}
\end{table*}

\methodname{} can be useful in detecting commonsense mistakes made by generative LMs in-the-wild.
We collected 27 anecdotes from the Internet where people reported \model{ChatGPT} making commonsense errors, and manually rewrote them into their correct versions, obtaining 54 statements in total.

% As reported in \autoref{tab:results_chatgpt_failures_metrics}, 
When detecting incorrect commonsense statements in this dataset, \methodname{} has a precision of 91\% and a recall of 74\%, amounting to an $F_1$ score of 82\%.
\autoref{tab:results_chatgpt_failures} shows how \methodname{} scores some of these these erroneous commonsense statements and their manually corrected version.
% We also manually corrected these incorrect statements and compare the scores received by both versions.
In 7 out of the 9 cases, \methodname{} assigns a low score to the original, incorrect statement, and a high score to the corrected statement.
For example, \textit{``since the density of a marble is much less than the density of mercury, the marble would sink to the bottom of the bowl if placed in it''} receives a score of 0.04 and is identified as an incorrect statement, whereas \textit{``since the density of a marble is much less than the density of mercury, the marble would
float if placed in mercury''} receives a score of 0.96 and is identified as a correct statement.
Meanwhile, there are also some failure cases.
\methodname{} believes that \textit{``it is possible for a solar eclipse to be followed by a lunar eclipse the next day''}, and fails to reject that \textit{``it is possible to draw a diagonal line in a triangle''}.

% \subsection{Further Analysis}
% \label{sec:further_analysis}
% We conduct ablation studies to investigate the effect of each training objective on seen and unseen (type 1) datasets. See \S\ref{sec:ablation} and \autoref{fig:results_ablations} (appendix) for full details. We also include the results when changing model sizes in \S\ref{sec:scaling} and \autoref{fig:results_scaling}. In this paper, we focus on the verification format to solve commonsense problems. A comprehensive discussion on how this format compares with the QA format is provided in \S\ref{sec:format} and \autoref{fig:results_format}.

\subsection{Analysis}
\label{sec:analysis}

\begin{figure*}[t]
\centering
\includegraphics[width=0.8\linewidth]{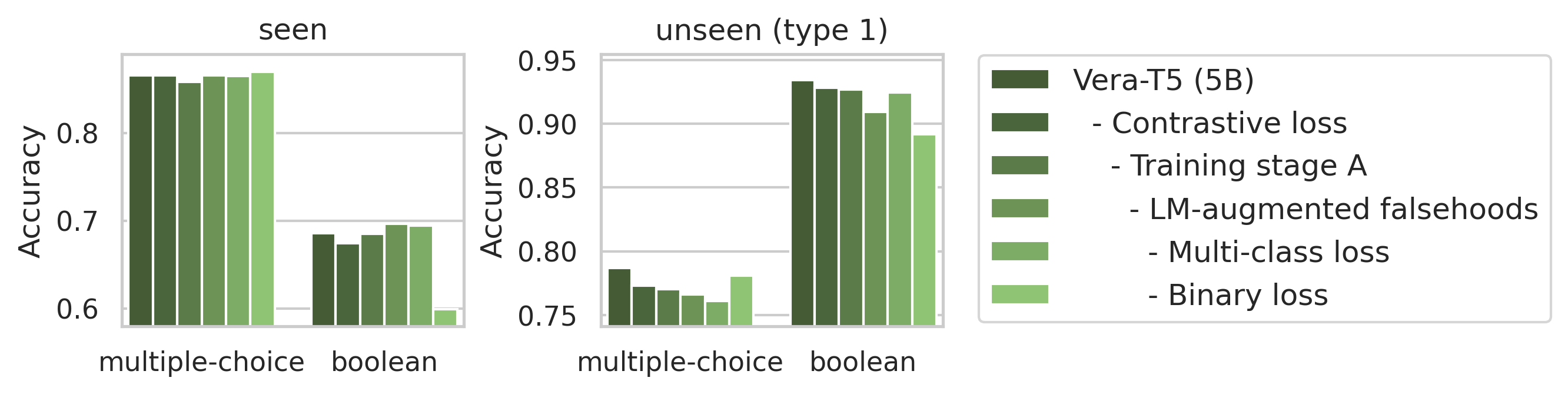}
\caption{
    Ablation results.
    Average accuracy on the development sets is reported.
    Components are incrementally removed from the training process, except for the multi-class loss and the binary loss; the hierarchy is indicated in the legend.
}
\label{fig:results_ablations}
% \vspace{-16pt}
\end{figure*}

\paragraph{Ablations.}
% \label{sec:ablation}
We conduct an ablation study by incrementally removing the following components from the training process: contrastive loss (\S\ref{sec:training_obj}), training stage A (\S\ref{sec:training_stages}), LM-augmented falsehoods (\S\ref{sec:data_construction}), multi-class loss or binary loss (\S\ref{sec:training_obj}).
Since at least one of the multi-class loss and the binary loss is needed, we remove them separately and observe the effect of training with a single loss.

\noindent
Results are shown in \autoref{fig:results_ablations}.
Overall, the ablated components have more impact on unseen benchmarks than seen ones.
Removing the contrastive loss hurts performance mostly on unseen datasets, implying that the contrastive objective is beneficial for generalization.
Removing training stage A hurts performance across the board, emphasizing the importance of training with large-scale commonsense knowledge.
LM-augmented falsehoods are most helpful on unseen benchmarks, with a little sacrifice in the performance on seen benchmarks.
The multi-class loss is most helpful on multiple-choice benchmarks, while removing the binary loss substantially hurts performance on boolean benchmarks. % because it causes the classification threshold to shift away from 0.

\begin{figure*}[t]
\centering
\includegraphics[width=0.8\linewidth]{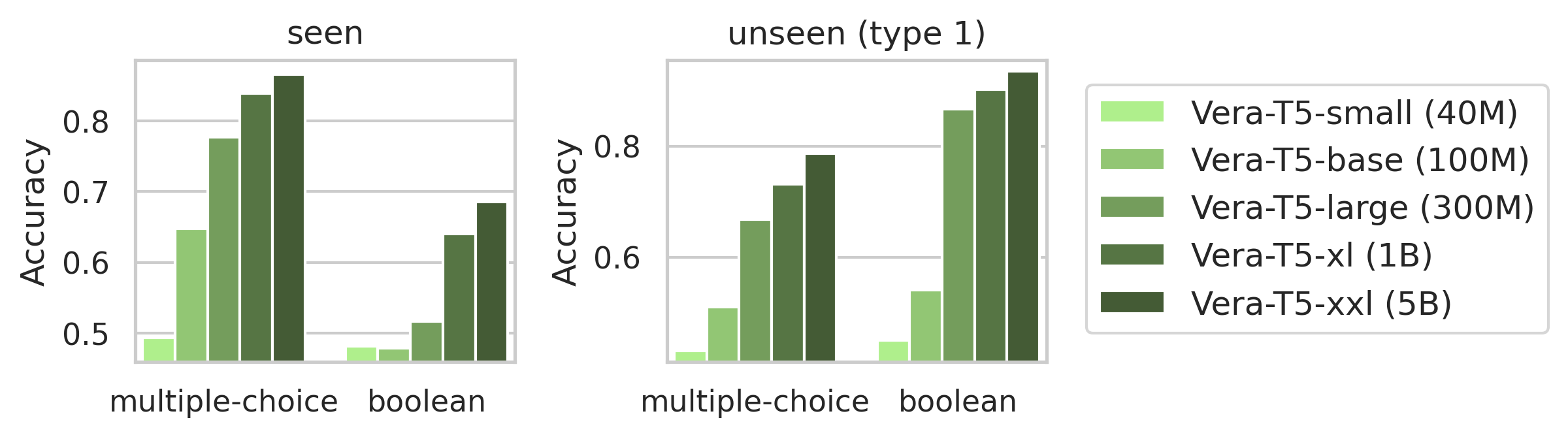}
\caption{
    Scaling trends of commonsense statement verifiers.
}
\label{fig:results_scaling}
% \vspace{-16pt}
\end{figure*}

\paragraph{Scaling Trends of \methodname{}.}
% \label{sec:scaling}
% We study the scaling trends of commonsense statement verifiers.
We trained variants of \methodname{} that are based on smaller versions of the \model{T5} encoder, and show the results in \autoref{fig:results_scaling}.
Model performance increases steadily with size, and does not show evidence of saturation at 5B parameters, suggesting that better commonsense plausibility estimation models might be yielded from larger pretrained LMs.

% \subsection{On-the-fly Commonsense Verification?}

\paragraph{Format: Verification vs. QA.}
In this paper, we focus on the verification format to solve commonsense problems. A comprehensive discussion on how this format compares with the QA format is provided in \S\ref{sec:further_analysis} and \autoref{fig:results_format}.

\section{Related Work}
\label{sec:related}

\paragraph{Commonsense verifiers.}
Prior work has explored the idea of verifying commonsense statements.
\textsc{Symbolic Knowledge Distillation} \citep{West2021SymbolicKD} and \textsc{I2D2} \citep{Bhagavatula2022I2D2IK} train models to classify the acceptability of model-generated commonsense statements.
The \textsc{Entailer} \citep{Tafjord2022EntailerAQ} model is partially trained to score the validity of a given hypothesis.
These models are trained on relatively small-scale, domain-specific data and do not generalize well to broader commonsense domains.
Some other work uses pretrained LMs with few-shot prompting to verify commonsense statements \citep{Kadavath2022LanguageM, Jung2022MaieuticPL}.
In this work, we develop a general-purpose commonsense statement verifier that works out-of-the-box in zero-shot setting.

\paragraph{Verification in other tasks.}
% Verification is an important methodology as it paves a path towards more robust and trustworthy AI systems.
Beyond commonsense statements, the problem of verification has been extensively studied on various NLP tasks.
NLI \citep{Liu2019RoBERTaAR, Liu2022WANLIWA, zhang2017ordinal} can be viewed as an \textit{entailment verification} task. % RoBERTa-MNLI, WANLI, JOCI
\citet{Chen2021CanNM} presents a method for \textit{QA verification} by transforming the context passage and question-answer pair into a premise-hypothesis format as in NLI.
Some work build models to perform \textit{reasoning verification} -- classifying whether a premise supports or refutes a hypothesis \citep{Bostrom2022NaturalLD, Sprague2022NaturalLD, Yang2022GeneratingNL, Tafjord2022EntailerAQ}. % SCSearch, ADGV, NLProofS, Entailer
On the other hand, \textit{fact verification} \citep{Thorne2018FEVERAL, Wadden2020FactOF} requires judging the validity of claims against a corpus of evidence (e.g., Wikipedia). % FEVER, SciFact
These tasks feature context-sensitive or knowledge-intensive hypotheses to verify and are typically complemented with additional context.
In contrast, we focus on verifying standalone commonsense statements where no context is required or provided.

\paragraph{Generation vs. verification.}
With the rapid progress in generative LMs, researchers have been largely building general-purpose problem-solving methods with a generative approach \citep{Khashabi2020UnifiedQACF, Khashabi2022UnifiedQAv2SG, Lourie2021UNICORNOR, Tafjord2021GeneralPurposeQW, Wei2022ChainOT}. % UnifiedQA, UnifiedQA-v2, UNICORN, Macaw, CoT
However, current generative LMs are still prone to hallucination errors and lack an intrinsic mechanism to express confidence level on their outputs.
Verification, on the other hand, shows promise to complement these shortcomings and has been adopted to improve the outcome of generation \citep{Chen2021CanNM, Jiang2022DraftSA}.
In this work, we take a pure verification approach and build a general-purpose verifier for commonsense statements, which to our best knowledge is the first of its kind.

\section{Conclusion and Future Work}
\label{sec:conclusion}

We introduced \methodname{}, a general-purpose verification model for commonsense statements and an early step toward tools for mitigating commonsense errors in text generated by language models. % \nascomment{restate the main findings/contributions.} \gary{Done}
\methodname{} achieves state-of-the-art performance when solving commonsense problems in the verification format, excels at filtering LM-generated commonsense knowledge statements, and is found useful in detecting erroneous commonsense statements from generative LMs. Furthermore, the scores produced by \methodname{} are well-calibrated; and could be used for plausibility score estimation for declarative statements if needed. As \methodname{} mainly targets on single-sentence statements, future work may consider verification of multi-sentence or long-form statements, or contextualized/defeasible commonsense statements.

% \nascomment{don't put this in conclusion; move to a limitations section} More work needs to be done to make models like \methodname{} deployable to real applications.
% For example, it needs a scope guard to determine whether the input text is a commonsense statement that falls into its capability to make a prediction.

% \nascomment{this doesn't go here, either.  I would just cut it} During experiments we found that standard RLHF \citep{Ouyang2022TrainingLM} might not be effective at endowing models with better commonsense abilities.
% Future work can consider using \methodname{} as a reward model to guide text generation.

\clearpage
\section*{Limitations}
\label{sec:limitations}

\methodname{} aims, and is trained, to predict the plausibility of statements based on objective commonsense knowledge of our world.
It is not intended to handle text outside the scope of commonsense statements (e.g., encyclopedic facts, reading comprehension with fictional worlds).
It is not trained or evaluated on moral commonsense data, so its capability of making moral predictions is unknown.
It gives a prediction even if the input falls out of its intended scope, which could be mitigated by an additional scope guard to determine its applicability.
In addition, it is not trained to handle very long and compositional input.
Although greatly outperforming existing systems, \methodname{} is not perfect and may make incorrect predictions.
It is not very robust under syntactic variations of the input, such as paraphrases and negations.
As the training data may contain bias or toxicity, \methodname{} may also make predictions that are perceived as ethically problematic.
The output of \methodname{} does not reflect the authors' view.
\methodname{} is a research prototype, and it is not designed for making real-world decisions.

\section*{Acknowledgments}
\label{sec:ack}

We thank Sean Welleck, Peter West, Alisa Liu, Jaehun Jung, Chandra Bhagavatula, Ram Pasunuru, Asli Celikyilmaz, and members of the H2lab, Xlab and ARK lab for their discussion and constructive feedback.
This work was funded in part by the DARPA MCS program through NIWC Pacific (N66001-19-2-4031), NSF IIS-2044660, and ONR N00014-18-1-2826.
We thank OpenAI for offering access to their API.

\bibliographystyle{acl_natbib}
\bibliography{custom}

\begin{thebibliography}{67}
\expandafter\ifx\csname natexlab\endcsname\relax\def\natexlab#1{#1}\fi

\bibitem[{Aroca-Ouellette et~al.(2021)Aroca-Ouellette, Paik, Roncone, and
  Kann}]{ArocaOuellette2021PROSTPR}
Stephane~T Aroca-Ouellette, Cory Paik, Alessandro Roncone, and Katharina Kann.
  2021.
\newblock Prost: Physical reasoning about objects through space and time.
\newblock In \emph{Findings}.

\bibitem[{Bhagavatula et~al.(2019)Bhagavatula, Bras, Malaviya, Sakaguchi,
  Holtzman, Rashkin, Downey, Yih, and Choi}]{Bhagavatula2019AbductiveCR}
Chandra Bhagavatula, Ronan~Le Bras, Chaitanya Malaviya, Keisuke Sakaguchi, Ari
  Holtzman, Hannah Rashkin, Doug Downey, Scott Yih, and Yejin Choi. 2019.
\newblock Abductive commonsense reasoning.
\newblock \emph{ArXiv}, abs/1908.05739.

\bibitem[{Bhagavatula et~al.(2022)Bhagavatula, Hwang, Downey, Bras, Lu,
  Sakaguchi, Swayamdipta, West, and Choi}]{Bhagavatula2022I2D2IK}
Chandra Bhagavatula, Jena~D. Hwang, Doug Downey, Ronan~Le Bras, Ximing Lu,
  Keisuke Sakaguchi, Swabha Swayamdipta, Peter West, and Yejin Choi. 2022.
\newblock I2d2: Inductive knowledge distillation with neurologic and
  self-imitation.
\newblock \emph{ArXiv}, abs/2212.09246.

\bibitem[{Bhakthavatsalam et~al.(2020)Bhakthavatsalam, Anastasiades, and
  Clark}]{Bhakthavatsalam2020GenericsKBAK}
Sumithra Bhakthavatsalam, Chloe Anastasiades, and Peter Clark. 2020.
\newblock Genericskb: A knowledge base of generic statements.
\newblock \emph{ArXiv}, abs/2005.00660.

\bibitem[{Bisk et~al.(2019)Bisk, Zellers, Bras, Gao, and Choi}]{Bisk2019PIQARA}
Yonatan Bisk, Rowan Zellers, Ronan~Le Bras, Jianfeng Gao, and Yejin Choi. 2019.
\newblock Piqa: Reasoning about physical commonsense in natural language.
\newblock \emph{ArXiv}, abs/1911.11641.

\bibitem[{Borji(2023)}]{Borji2023ACA}
Ali Borji. 2023.
\newblock A categorical archive of chatgpt failures.
\newblock \emph{ArXiv}, abs/2302.03494.

\bibitem[{Bostrom et~al.(2022)Bostrom, Sprague, Chaudhuri, and
  Durrett}]{Bostrom2022NaturalLD}
Kaj Bostrom, Zayne Sprague, Swarat Chaudhuri, and Greg Durrett. 2022.
\newblock Natural language deduction through search over statement
  compositions.
\newblock In \emph{Conference on Empirical Methods in Natural Language
  Processing}.

\bibitem[{Brown et~al.(2020)Brown, Mann, Ryder, Subbiah, Kaplan, Dhariwal,
  Neelakantan, Shyam, Sastry, Askell, Agarwal, Herbert-Voss, Krueger, Henighan,
  Child, Ramesh, Ziegler, Wu, Winter, Hesse, Chen, Sigler, Litwin, Gray, Chess,
  Clark, Berner, McCandlish, Radford, Sutskever, and
  Amodei}]{Brown2020LanguageMA}
Tom~B. Brown, Benjamin Mann, Nick Ryder, Melanie Subbiah, Jared Kaplan,
  Prafulla Dhariwal, Arvind Neelakantan, Pranav Shyam, Girish Sastry, Amanda
  Askell, Sandhini Agarwal, Ariel Herbert-Voss, Gretchen Krueger, T.~J.
  Henighan, Rewon Child, Aditya Ramesh, Daniel~M. Ziegler, Jeff Wu, Clemens
  Winter, Christopher Hesse, Mark Chen, Eric Sigler, Mateusz Litwin, Scott
  Gray, Benjamin Chess, Jack Clark, Christopher Berner, Sam McCandlish, Alec
  Radford, Ilya Sutskever, and Dario Amodei. 2020.
\newblock Language models are few-shot learners.
\newblock \emph{ArXiv}, abs/2005.14165.

\bibitem[{Chen et~al.(2021)Chen, Choi, and Durrett}]{Chen2021CanNM}
Jifan Chen, Eunsol Choi, and Greg Durrett. 2021.
\newblock Can nli models verify qa systems' predictions?
\newblock \emph{ArXiv}, abs/2104.08731.

\bibitem[{Chen et~al.(2019)Chen, D'Arcy, Liu, Fernandez, and
  Downey}]{chen2019codah}
Michael Chen, Mike D'Arcy, Alisa Liu, Jared Fernandez, and Doug Downey. 2019.
\newblock Codah: An adversarially authored question-answer dataset for common
  sense.
\newblock \emph{arXiv preprint arXiv:1904.04365}.

\bibitem[{Chung et~al.(2022)Chung, Hou, Longpre, Zoph, Tay, Fedus, Li, Wang,
  Dehghani, Brahma, Webson, Gu, Dai, Suzgun, Chen, Chowdhery, Valter, Narang,
  Mishra, Yu, Zhao, Huang, Dai, Yu, Petrov, hsin Chi, Dean, Devlin, Roberts,
  Zhou, Le, and Wei}]{Chung2022ScalingIL}
Hyung~Won Chung, Le~Hou, S.~Longpre, Barret Zoph, Yi~Tay, William Fedus, Eric
  Li, Xuezhi Wang, Mostafa Dehghani, Siddhartha Brahma, Albert Webson,
  Shixiang~Shane Gu, Zhuyun Dai, Mirac Suzgun, Xinyun Chen, Aakanksha
  Chowdhery, Dasha Valter, Sharan Narang, Gaurav Mishra, Adams~Wei Yu, Vincent
  Zhao, Yanping Huang, Andrew~M. Dai, Hongkun Yu, Slav Petrov, Ed~Huai hsin
  Chi, Jeff Dean, Jacob Devlin, Adam Roberts, Denny Zhou, Quoc~V. Le, and Jason
  Wei. 2022.
\newblock Scaling instruction-finetuned language models.
\newblock \emph{ArXiv}, abs/2210.11416.

\bibitem[{Clark et~al.(2018)Clark, Cowhey, Etzioni, Khot, Sabharwal, Schoenick,
  and Tafjord}]{Clark2018ThinkYH}
Peter Clark, Isaac Cowhey, Oren Etzioni, Tushar Khot, Ashish Sabharwal, Carissa
  Schoenick, and Oyvind Tafjord. 2018.
\newblock Think you have solved question answering? try arc, the ai2 reasoning
  challenge.
\newblock \emph{ArXiv}, abs/1803.05457.

\bibitem[{Davis(2023)}]{Davis2023BenchmarksFA}
Ernest Davis. 2023.
\newblock Benchmarks for automated commonsense reasoning: A survey.
\newblock \emph{ArXiv}, abs/2302.04752.

\bibitem[{Geva et~al.(2021)Geva, Khashabi, Segal, Khot, Roth, and
  Berant}]{Geva2021DidAU}
Mor Geva, Daniel Khashabi, Elad Segal, Tushar Khot, Dan Roth, and Jonathan
  Berant. 2021.
\newblock Did aristotle use a laptop? a question answering benchmark with
  implicit reasoning strategies.
\newblock \emph{Transactions of the Association for Computational Linguistics},
  9:346--361.

\bibitem[{Gordon et~al.(2011)Gordon, Kozareva, and
  Roemmele}]{Gordon2011SemEval2012T7}
Andrew~S. Gordon, Zornitsa Kozareva, and Melissa Roemmele. 2011.
\newblock Semeval-2012 task 7: Choice of plausible alternatives: An evaluation
  of commonsense causal reasoning.
\newblock In \emph{International Workshop on Semantic Evaluation}.

\bibitem[{Gugger et~al.(2022)Gugger, Debut, Wolf, Schmid, Mueller, and
  Mangrulkar}]{accelerate}
Sylvain Gugger, Lysandre Debut, Thomas Wolf, Philipp Schmid, Zachary Mueller,
  and Sourab Mangrulkar. 2022.
\newblock Accelerate: Training and inference at scale made simple, efficient
  and adaptable.
\newblock \url{https://github.com/huggingface/accelerate}.

\bibitem[{Guo et~al.(2017)Guo, Pleiss, Sun, and Weinberger}]{Guo2017OnCO}
Chuan Guo, Geoff Pleiss, Yu~Sun, and Kilian~Q. Weinberger. 2017.
\newblock On calibration of modern neural networks.
\newblock In \emph{International Conference on Machine Learning}.

\bibitem[{Hwang et~al.(2020)Hwang, Bhagavatula, Bras, Da, Sakaguchi, Bosselut,
  and Choi}]{Hwang2020COMETATOMIC2O}
Jena~D. Hwang, Chandra Bhagavatula, Ronan~Le Bras, Jeff Da, Keisuke Sakaguchi,
  Antoine Bosselut, and Yejin Choi. 2020.
\newblock Comet-atomic 2020: On symbolic and neural commonsense knowledge
  graphs.
\newblock In \emph{AAAI Conference on Artificial Intelligence}.

\bibitem[{Jiang et~al.(2022)Jiang, Welleck, Zhou, Li, Liu, Jamnik, Lacroix, Wu,
  and Lample}]{Jiang2022DraftSA}
Albert~Qiaochu Jiang, Sean Welleck, Jin~Peng Zhou, Wenda Li, Jiacheng Liu,
  Mateja Jamnik, Timoth{\'e}e Lacroix, Yuhuai Wu, and Guillaume Lample. 2022.
\newblock Draft, sketch, and prove: Guiding formal theorem provers with
  informal proofs.
\newblock \emph{ArXiv}, abs/2210.12283.

\bibitem[{Jung et~al.(2022)Jung, Qin, Welleck, Brahman, Bhagavatula, Bras, and
  Choi}]{Jung2022MaieuticPL}
Jaehun Jung, Lianhui Qin, Sean Welleck, Faeze Brahman, Chandra Bhagavatula,
  Ronan~Le Bras, and Yejin Choi. 2022.
\newblock Maieutic prompting: Logically consistent reasoning with recursive
  explanations.
\newblock In \emph{Conference on Empirical Methods in Natural Language
  Processing}.

\bibitem[{Kadavath et~al.(2022)Kadavath, Conerly, Askell, Henighan, Drain,
  Perez, Schiefer, Dodds, DasSarma, Tran-Johnson, Johnston, El-Showk, Jones,
  Elhage, Hume, Chen, Bai, Bowman, Fort, Ganguli, Hernandez, Jacobson, Kernion,
  Kravec, Lovitt, Ndousse, Olsson, Ringer, Amodei, Brown, Clark, Joseph, Mann,
  McCandlish, Olah, and Kaplan}]{Kadavath2022LanguageM}
Saurav Kadavath, Tom Conerly, Amanda Askell, T.~J. Henighan, Dawn Drain, Ethan
  Perez, Nicholas Schiefer, Zachary Dodds, Nova DasSarma, Eli Tran-Johnson,
  Scott Johnston, Sheer El-Showk, Andy Jones, Nelson Elhage, Tristan Hume, Anna
  Chen, Yuntao Bai, Sam Bowman, Stanislav Fort, Deep Ganguli, Danny Hernandez,
  Josh Jacobson, John Kernion, Shauna Kravec, Liane Lovitt, Kamal Ndousse,
  Catherine Olsson, Sam Ringer, Dario Amodei, Tom~B. Brown, Jack Clark,
  Nicholas Joseph, Benjamin Mann, Sam McCandlish, Christopher Olah, and Jared
  Kaplan. 2022.
\newblock Language models (mostly) know what they know.
\newblock \emph{ArXiv}, abs/2207.05221.

\bibitem[{Khashabi et~al.(2022)Khashabi, Kordi, and
  Hajishirzi}]{Khashabi2022UnifiedQAv2SG}
Daniel Khashabi, Yeganeh Kordi, and Hannaneh Hajishirzi. 2022.
\newblock Unifiedqa-v2: Stronger generalization via broader cross-format
  training.
\newblock \emph{ArXiv}, abs/2202.12359.

\bibitem[{Khashabi et~al.(2020)Khashabi, Min, Khot, Sabharwal, Tafjord, Clark,
  and Hajishirzi}]{Khashabi2020UnifiedQACF}
Daniel Khashabi, Sewon Min, Tushar Khot, Ashish Sabharwal, Oyvind Tafjord,
  Peter Clark, and Hannaneh Hajishirzi. 2020.
\newblock Unifiedqa: Crossing format boundaries with a single qa system.
\newblock In \emph{Findings}.

\bibitem[{Khosla et~al.(2020)Khosla, Teterwak, Wang, Sarna, Tian, Isola,
  Maschinot, Liu, and Krishnan}]{Khosla2020SuperCL}
Prannay Khosla, Piotr Teterwak, Chen Wang, Aaron Sarna, Yonglong Tian, Phillip
  Isola, Aaron Maschinot, Ce~Liu, and Dilip Krishnan. 2020.
\newblock Supervised contrastive learning.
\newblock In \emph{Advances in Neural Information Processing Systems},
  volume~33.

\bibitem[{Khot et~al.(2019)Khot, Clark, Guerquin, Jansen, and
  Sabharwal}]{Khot2019QASCAD}
Tushar Khot, Peter Clark, Michal Guerquin, Peter~Alexander Jansen, and Ashish
  Sabharwal. 2019.
\newblock Qasc: A dataset for question answering via sentence composition.
\newblock \emph{ArXiv}, abs/1910.11473.

\bibitem[{Kingma and Ba(2014)}]{Kingma2014AdamAM}
Diederik~P. Kingma and Jimmy Ba. 2014.
\newblock Adam: A method for stochastic optimization.
\newblock \emph{CoRR}, abs/1412.6980.

\bibitem[{Levesque et~al.(2011)Levesque, Davis, and
  Morgenstern}]{Levesque2011TheWS}
Hector~J. Levesque, Ernest Davis, and L.~Morgenstern. 2011.
\newblock The winograd schema challenge.
\newblock In \emph{International Conference on Principles of Knowledge
  Representation and Reasoning}.

\bibitem[{Lin et~al.(2020)Lin, Lee, Khanna, and Ren}]{Lin2020BirdsHF}
Bill~Yuchen Lin, Seyeon Lee, Rahul Khanna, and Xiang Ren. 2020.
\newblock Birds have four legs?! numersense: Probing numerical commonsense
  knowledge of pre-trained language models.
\newblock \emph{ArXiv}, abs/2005.00683.

\bibitem[{Lin et~al.(2022)Lin, Hilton, and Evans}]{lin-etal-2022-truthfulqa}
Stephanie Lin, Jacob Hilton, and Owain Evans. 2022.
\newblock {T}ruthful{QA}: Measuring how models mimic human falsehoods.
\newblock In \emph{ACL}, pages 3214--3252.

\bibitem[{Liu et~al.(2022{\natexlab{a}})Liu, Swayamdipta, Smith, and
  Choi}]{Liu2022WANLIWA}
Alisa Liu, Swabha Swayamdipta, Noah~A. Smith, and Yejin Choi.
  2022{\natexlab{a}}.
\newblock Wanli: Worker and ai collaboration for natural language inference
  dataset creation.
\newblock In \emph{Conference on Empirical Methods in Natural Language
  Processing}.

\bibitem[{Liu et~al.(2022{\natexlab{b}})Liu, Hallinan, Lu, He, Welleck,
  Hajishirzi, and Choi}]{Liu2022RainierRK}
Jiacheng Liu, Skyler Hallinan, Ximing Lu, Pengfei He, Sean Welleck, Hannaneh
  Hajishirzi, and Yejin Choi. 2022{\natexlab{b}}.
\newblock Rainier: Reinforced knowledge introspector for commonsense question
  answering.
\newblock In \emph{Conference on Empirical Methods in Natural Language
  Processing}.

\bibitem[{Liu et~al.(2021)Liu, Liu, Lu, Welleck, West, Bras, Choi, and
  Hajishirzi}]{Liu2021GeneratedKP}
Jiacheng Liu, Alisa Liu, Ximing Lu, Sean Welleck, Peter West, Ronan~Le Bras,
  Yejin Choi, and Hannaneh Hajishirzi. 2021.
\newblock Generated knowledge prompting for commonsense reasoning.
\newblock \emph{ArXiv}, abs/2110.08387.

\bibitem[{Liu et~al.(2022{\natexlab{c}})Liu, Yin, Feng, and
  Zhao}]{Liu2022ThingsNW}
Xiao Liu, Da~Yin, Yansong Feng, and Dongyan Zhao. 2022{\natexlab{c}}.
\newblock Things not written in text: Exploring spatial commonsense from visual
  signals.
\newblock \emph{ArXiv}, abs/2203.08075.

\bibitem[{Liu et~al.(2019)Liu, Ott, Goyal, Du, Joshi, Chen, Levy, Lewis,
  Zettlemoyer, and Stoyanov}]{Liu2019RoBERTaAR}
Yinhan Liu, Myle Ott, Naman Goyal, Jingfei Du, Mandar Joshi, Danqi Chen, Omer
  Levy, Mike Lewis, Luke Zettlemoyer, and Veselin Stoyanov. 2019.
\newblock Roberta: A robustly optimized bert pretraining approach.
\newblock \emph{ArXiv}, abs/1907.11692.

\bibitem[{Lourie et~al.(2021)Lourie, Bras, Bhagavatula, and
  Choi}]{Lourie2021UNICORNOR}
Nicholas Lourie, Ronan~Le Bras, Chandra Bhagavatula, and Yejin Choi. 2021.
\newblock Unicorn on rainbow: A universal commonsense reasoning model on a new
  multitask benchmark.
\newblock In \emph{AAAI Conference on Artificial Intelligence}.

\bibitem[{Marcus and
  Davis(2023)}]{1kDSERnROv5FgHbVN8z_bXH9gak2IXRtoqz0nwhrviCw}
Gary Marcus and Ernest Davis. 2023.
\newblock Chatgpt/llm errors (public).
\newblock
  \url{https://docs.google.com/spreadsheets/d/1kDSERnROv5FgHbVN8z_bXH9gak2IXRtoqz0nwhrviCw/edit}.

\bibitem[{Mihaylov et~al.(2018)Mihaylov, Clark, Khot, and
  Sabharwal}]{Mihaylov2018CanAS}
Todor Mihaylov, Peter Clark, Tushar Khot, and Ashish Sabharwal. 2018.
\newblock Can a suit of armor conduct electricity? a new dataset for open book
  question answering.
\newblock In \emph{Conference on Empirical Methods in Natural Language
  Processing}.

\bibitem[{Mostafazadeh et~al.(2016)Mostafazadeh, Chambers, He, Parikh, Batra,
  Vanderwende, Kohli, and Allen}]{Mostafazadeh2016ACA}
N.~Mostafazadeh, Nathanael Chambers, Xiaodong He, Devi Parikh, Dhruv Batra,
  Lucy Vanderwende, Pushmeet Kohli, and James~F. Allen. 2016.
\newblock A corpus and cloze evaluation for deeper understanding of commonsense
  stories.
\newblock In \emph{North American Chapter of the Association for Computational
  Linguistics}.

\bibitem[{Naeini et~al.(2015)Naeini, Cooper, and
  Hauskrecht}]{Naeini2015ObtainingWC}
Mahdi~Pakdaman Naeini, Gregory~F. Cooper, and Milos Hauskrecht. 2015.
\newblock Obtaining well calibrated probabilities using bayesian binning.
\newblock \emph{Proceedings of the ... AAAI Conference on Artificial
  Intelligence. AAAI Conference on Artificial Intelligence}, 2015:2901--2907.

\bibitem[{Onoe et~al.(2021)Onoe, Zhang, Choi, and Durrett}]{Onoe2021CREAKAD}
Yasumasa Onoe, Michael~J.Q. Zhang, Eunsol Choi, and Greg Durrett. 2021.
\newblock Creak: A dataset for commonsense reasoning over entity knowledge.
\newblock \emph{ArXiv}, abs/2109.01653.

\bibitem[{OpenAI(2022{\natexlab{a}})}]{chatgptblog}
OpenAI. 2022{\natexlab{a}}.
\newblock \href {https://openai.com/blog/chatgpt} {Introducing chatgpt}.

\bibitem[{OpenAI(2022{\natexlab{b}})}]{gpt35docs}
OpenAI. 2022{\natexlab{b}}.
\newblock \href {https://platform.openai.com/docs/models/gpt-3-5} {Models -
  overview - gpt-3.5}.

\bibitem[{OpenAI(2023)}]{gpt4}
OpenAI. 2023.
\newblock \href {https://cdn.openai.com/papers/gpt-4.pdf} {Gpt-4 technical
  report}.
\newblock Technical report, OpenAI.

\bibitem[{Raffel et~al.(2020)Raffel, Shazeer, Roberts, Lee, Narang, Matena,
  Zhou, Li, and Liu}]{Raffel2020t5}
Colin Raffel, Noam Shazeer, Adam Roberts, Katherine Lee, Sharan Narang, Michael
  Matena, Yanqi Zhou, Wei Li, and Peter~J. Liu. 2020.
\newblock \href {http://jmlr.org/papers/v21/20-074.html} {Exploring the limits
  of transfer learning with a unified text-to-text transformer}.
\newblock \emph{Journal of Machine Learning Research}, 21(140):1--67.

\bibitem[{Sakaguchi et~al.(2019)Sakaguchi, Bras, Bhagavatula, and
  Choi}]{Sakaguchi2019WINOGRANDEAA}
Keisuke Sakaguchi, Ronan~Le Bras, Chandra Bhagavatula, and Yejin Choi. 2019.
\newblock Winogrande: An adversarial winograd schema challenge at scale.
\newblock \emph{Commun. ACM}, 64:99--106.

\bibitem[{Sap et~al.(2019)Sap, Rashkin, Chen, Bras, and Choi}]{Sap2019SocialIC}
Maarten Sap, Hannah Rashkin, Derek Chen, Ronan~Le Bras, and Yejin Choi. 2019.
\newblock Social iqa: Commonsense reasoning about social interactions.
\newblock \emph{ArXiv}, abs/1904.09728.

\bibitem[{Singh et~al.(2021)Singh, Wen, Hou, Alipoormolabashi, Wu, Ma, and
  Peng}]{Singh2021COM2SENSEAC}
Shikhar Singh, Nuan Wen, Yu~Hou, Pegah Alipoormolabashi, Te-Lin Wu, Xuezhe Ma,
  and Nanyun Peng. 2021.
\newblock Com2sense: A commonsense reasoning benchmark with complementary
  sentences.
\newblock In \emph{Findings}.

\bibitem[{Sprague et~al.(2022)Sprague, Bostrom, Chaudhuri, and
  Durrett}]{Sprague2022NaturalLD}
Zayne Sprague, Kaj Bostrom, Swarat Chaudhuri, and Greg Durrett. 2022.
\newblock Natural language deduction with incomplete information.
\newblock In \emph{Conference on Empirical Methods in Natural Language
  Processing}.

\bibitem[{Tafjord and Clark(2021)}]{Tafjord2021GeneralPurposeQW}
Oyvind Tafjord and Peter Clark. 2021.
\newblock General-purpose question-answering with macaw.
\newblock \emph{ArXiv}, abs/2109.02593.

\bibitem[{Tafjord et~al.(2018)Tafjord, Clark, Gardner, tau Yih, and
  Sabharwal}]{Tafjord2018QuaRelAD}
Oyvind Tafjord, Peter Clark, Matt Gardner, Wen tau Yih, and Ashish Sabharwal.
  2018.
\newblock Quarel: A dataset and models for answering questions about
  qualitative relationships.
\newblock \emph{ArXiv}, abs/1811.08048.

\bibitem[{Tafjord et~al.(2022)Tafjord, Dalvi, and
  Clark}]{Tafjord2022EntailerAQ}
Oyvind Tafjord, Bhavana Dalvi, and Peter Clark. 2022.
\newblock Entailer: Answering questions with faithful and truthful chains of
  reasoning.
\newblock In \emph{Conference on Empirical Methods in Natural Language
  Processing}.

\bibitem[{Tafjord et~al.(2019)Tafjord, Gardner, Lin, and
  Clark}]{Tafjord2019QuaRTzAO}
Oyvind Tafjord, Matt Gardner, Kevin Lin, and Peter Clark. 2019.
\newblock Quartz: An open-domain dataset of qualitative relationship questions.
\newblock \emph{ArXiv}, abs/1909.03553.

\bibitem[{Talmor et~al.(2019)Talmor, Herzig, Lourie, and
  Berant}]{Talmor2019CommonsenseQAAQ}
Alon Talmor, Jonathan Herzig, Nicholas Lourie, and Jonathan Berant. 2019.
\newblock Commonsenseqa: A question answering challenge targeting commonsense
  knowledge.
\newblock \emph{ArXiv}, abs/1811.00937.

\bibitem[{Talmor et~al.(2021)Talmor, Yoran, Bras, Bhagavatula, Goldberg, Choi,
  and Berant}]{Talmor2021CommonsenseQA2E}
Alon Talmor, Ori Yoran, Ronan~Le Bras, Chandrasekhar Bhagavatula, Yoav
  Goldberg, Yejin Choi, and Jonathan Berant. 2021.
\newblock Commonsenseqa 2.0: Exposing the limits of ai through gamification.
\newblock \emph{ArXiv}, abs/2201.05320.

\bibitem[{Thorne et~al.(2018)Thorne, Vlachos, Christodoulopoulos, and
  Mittal}]{Thorne2018FEVERAL}
James Thorne, Andreas Vlachos, Christos Christodoulopoulos, and Arpit Mittal.
  2018.
\newblock Fever: a large-scale dataset for fact extraction and verification.
\newblock \emph{ArXiv}, abs/1803.05355.

\bibitem[{Touvron et~al.(2023)Touvron, Lavril, Izacard, Martinet, Lachaux,
  Lacroix, Rozi{\`e}re, Goyal, Hambro, Azhar, Rodriguez, Joulin, Grave, and
  Lample}]{Touvron2023LLaMAOA}
Hugo Touvron, Thibaut Lavril, Gautier Izacard, Xavier Martinet, Marie-Anne
  Lachaux, Timoth{\'e}e Lacroix, Baptiste Rozi{\`e}re, Naman Goyal, Eric
  Hambro, Faisal Azhar, Aur'elien Rodriguez, Armand Joulin, Edouard Grave, and
  Guillaume Lample. 2023.
\newblock Llama: Open and efficient foundation language models.
\newblock \emph{ArXiv}, abs/2302.13971.

\bibitem[{Venuto(2023)}]{giuven95/chatgpt-failures}
Giuseppe Venuto. 2023.
\newblock chatgpt-failures.
\newblock \url{https://github.com/giuven95/chatgpt-failures}.

\bibitem[{Wadden et~al.(2020)Wadden, Lo, Wang, Lin, van Zuylen, Cohan, and
  Hajishirzi}]{Wadden2020FactOF}
David Wadden, Kyle Lo, Lucy~Lu Wang, Shanchuan Lin, Madeleine van Zuylen, Arman
  Cohan, and Hannaneh Hajishirzi. 2020.
\newblock Fact or fiction: Verifying scientific claims.
\newblock \emph{ArXiv}, abs/2004.14974.

\bibitem[{Wang et~al.(2020)Wang, Liang, Jin, Wang, Zhu, and
  Zhang}]{Wang2020SemEval2020T4}
Cunxiang Wang, Shuailong Liang, Yili Jin, Yilong Wang, Xiaodan Zhu, and Yue
  Zhang. 2020.
\newblock Semeval-2020 task 4: Commonsense validation and explanation.
\newblock In \emph{International Workshop on Semantic Evaluation}.

\bibitem[{Wei et~al.(2022)Wei, Wang, Schuurmans, Bosma, hsin Chi, Le, and
  Zhou}]{Wei2022ChainOT}
Jason Wei, Xuezhi Wang, Dale Schuurmans, Maarten Bosma, Ed~Huai hsin Chi, Quoc
  Le, and Denny Zhou. 2022.
\newblock Chain of thought prompting elicits reasoning in large language
  models.
\newblock \emph{ArXiv}, abs/2201.11903.

\bibitem[{Welbl et~al.(2017)Welbl, Liu, and Gardner}]{Welbl2017CrowdsourcingMC}
Johannes Welbl, Nelson~F. Liu, and Matt Gardner. 2017.
\newblock Crowdsourcing multiple choice science questions.
\newblock \emph{ArXiv}, abs/1707.06209.

\bibitem[{West et~al.(2021)West, Bhagavatula, Hessel, Hwang, Jiang, Bras, Lu,
  Welleck, and Choi}]{West2021SymbolicKD}
Peter West, Chandrasekhar Bhagavatula, Jack Hessel, Jena~D. Hwang, Liwei Jiang,
  Ronan~Le Bras, Ximing Lu, Sean Welleck, and Yejin Choi. 2021.
\newblock Symbolic knowledge distillation: from general language models to
  commonsense models.
\newblock In \emph{North American Chapter of the Association for Computational
  Linguistics}.

\bibitem[{Wolf et~al.(2019)Wolf, Debut, Sanh, Chaumond, Delangue, Moi, Cistac,
  Rault, Louf, Funtowicz, and Brew}]{Wolf2019HuggingFacesTS}
Thomas Wolf, Lysandre Debut, Victor Sanh, Julien Chaumond, Clement Delangue,
  Anthony Moi, Pierric Cistac, Tim Rault, R{\'e}mi Louf, Morgan Funtowicz, and
  Jamie Brew. 2019.
\newblock Huggingface's transformers: State-of-the-art natural language
  processing.
\newblock \emph{ArXiv}, abs/1910.03771.

\bibitem[{Yang et~al.(2022)Yang, Deng, and Chen}]{Yang2022GeneratingNL}
Kaiyu Yang, Jia Deng, and Danqi Chen. 2022.
\newblock Generating natural language proofs with verifier-guided search.
\newblock In \emph{Conference on Empirical Methods in Natural Language
  Processing}.

\bibitem[{Zellers et~al.(2018)Zellers, Bisk, Schwartz, and
  Choi}]{Zellers2018SWAGAL}
Rowan Zellers, Yonatan Bisk, Roy Schwartz, and Yejin Choi. 2018.
\newblock Swag: A large-scale adversarial dataset for grounded commonsense
  inference.
\newblock In \emph{Conference on Empirical Methods in Natural Language
  Processing}.

\bibitem[{Zellers et~al.(2019)Zellers, Holtzman, Bisk, Farhadi, and
  Choi}]{Zellers2019HellaSwagCA}
Rowan Zellers, Ari Holtzman, Yonatan Bisk, Ali Farhadi, and Yejin Choi. 2019.
\newblock Hellaswag: Can a machine really finish your sentence?
\newblock In \emph{Annual Meeting of the Association for Computational
  Linguistics}.

\bibitem[{Zhang et~al.(2017)Zhang, Rudinger, Duh, and
  Van~Durme}]{zhang2017ordinal}
Sheng Zhang, Rachel Rudinger, Kevin Duh, and Benjamin Van~Durme. 2017.
\newblock Ordinal common-sense inference.
\newblock \emph{Transactions of the Association for Computational Linguistics},
  5:379--395.

\end{thebibliography}

\clearpage
\appendix

\section{More Details on Datasets}
\label{sec:datasets_more}

\autoref{tab:datasets_stat} shows more dataset statistics, and \autoref{tab:datasets_more} shows the dataset citations and links from which we retrieved the datasets.

\subsection{Dataset-Specific Special Handling}

For some datasets, we pre-process them into a unified multiple-choice or boolean format.
We provide the details below.

\paragraph{\ds{Com2Sense (paired)}.}
\ds{Com2Sense} contains true and false statements that can be paired into complements.
To utilize this pairing information, we place the two statements in each pair into the same statement group, and treat this as a multiple-choice dataset.
Some statements in the dev set are not paired, so we discarded these examples.

\paragraph{\ds{CycIC (mc)}.}
\ds{CycIC} contains both multiple-choice and boolean QA problems.
To keep consistency in evaluation, we use only the multiple-choice problems, which is the dominant problem type in this dataset.

\paragraph{\ds{ComVE (task A)}.}
\ds{ComVE} contains data for three tasks.
Task A is assigning true/false labels to paired statements, similar to \ds{Com2Sense (paired)}.
Task B and C are about choosing and generating explanations to a given statement being against commonsense.
We use the data for task A.

\paragraph{\ds{SKD (annotated)}.}
The annotated dataset of Symbolic Knowledge Distillation (\ds{SKD}) contains LM-generated, semi-structured knowledge triples, where the head and tail events are connected by relations, such as
\begin{quote}
\texttt{\small (PersonX doesn't like to wait, xIntent, to get the job done)}.
\end{quote}
Following \citet{West2021SymbolicKD}, we replace the name placeholders with random person names, and convert into natural language statements using templates adapted from \citet{Hwang2020COMETATOMIC2O}.
For example, the triple in the above example becomes
\begin{quote}
\texttt{\small Arnold doesn't like to wait. Because Arnold wanted to get the job done.}
\end{quote}
We set the correctness label to be true iff the \textit{valid} field has a positive value.

\paragraph{\ds{I2D2 (annotated)}.}
The annotated dataset of I2D2 contains LM-generated commonsense statements with human-annotated correctness labels.
We use the combination of annotated data in ``Iter0'' and ``Iter2'', because the data of ``Iter1'' is missing from the website.

\subsection{Conversion to Declarative Statements}
\label{sec:conversion}

From QA datasets, we create declarative statements from QA problems using the following method:
\begin{itemize}
\item If the problem contains a question, we convert the question and choice into a declarative statement using the question conversion model created by \citet{Chen2021CanNM}.
\item If the question is cloze-style, we replace the blank with the choice.
\item If the question is an incomplete sentence and the choice is a continuation to it, we concatenate the question and the choice.
\item If there is no question and the problem only asks to choose between some choices, we use the choice as the declarative statement.
\item For boolean problems, we always use \textit{yes} as the choice and create a single declarative statement for each problem. We use the original label as the correctness label of this statement.
\end{itemize}

% \subsection{LM-augmented falsehoods}

\begin{table*}[t]
\small
\setlength{\tabcolsep}{3pt}
\centering
\resizebox{\textwidth}{!}{%
\begin{tabular}{l l c c r c r r r r}
\toprule
\textbf{Abbr.} & \textbf{Name} & \textbf{Domain} & \textbf{Format} & \textbf{\# Train Ex.} & \textbf{Aug} & \textbf{\# Dev Ex.} & \textbf{\# Statements} & \textbf{\# True} & \textbf{\# False} \\
\midrule
\multicolumn{10}{c}{\textsc{Stage A training}} \\
\midrule
% The following content is automatically generated by `python dsstat.py`
Atomic2020    & Atomic2020               &              & multiple-choice (4)  & 803541 &        &  70731 & 282924 &  70731 & 212193 \\
GenericsKB    & GenericsKB               &              & multiple-choice (4)  & 775820 &        &  96977 & 387908 &  96977 & 290931 \\
\textbf{Total} & & & & \textbf{1579361} & & \textbf{167708} & \textbf{670832} & \textbf{167708} & \textbf{503124} \\
\midrule
\multicolumn{10}{c}{\textsc{Stage B training (seen)}} \\
\midrule
OBQA          & OpenBookQA               & scientific   & multiple-choice (4)  &   4957 & \cmark &    500 &   2000 &    500 &   1500 \\
ARC\_e        & ARC (easy)               & scientific   & multiple-choice (4)  &   2251 & \cmark &    570 &   2281 &    570 &   1711 \\
ARC\_h        & ARC (hard)               & scientific   & multiple-choice (4)  &   1119 & \cmark &    299 &   1194 &    299 &    895 \\
AI2Sci\_e     & AI2 Science (elem)       & scientific   & multiple-choice (4)  &    623 & \cmark &    123 &    489 &    123 &    366 \\
AI2Sci\_m     & AI2 Science (middle)     & scientific   & multiple-choice (4)  &    605 & \cmark &    125 &    502 &    125 &    377 \\
CSQA          & CommonsenseQA            & general      & multiple-choice (5)  &   9741 & \cmark &   1221 &   6099 &   1221 &   4878 \\
QASC          & QASC                     & scientific   & multiple-choice (8)  &   8134 & \cmark &    926 &   7408 &    926 &   6482 \\
PIQA          & Physical IQA             & physical     & multiple-choice (2)  &  16113 &        &   1838 &   3676 &   1838 &   1838 \\
SIQA          & Social IQA               & social       & multiple-choice (3)  &  33410 & \cmark &   1954 &   5861 &   1954 &   3907 \\
WG            & Winogrande               & general      & multiple-choice (2)  &  40398 &        &   1267 &   2534 &   1267 &   1267 \\
C2S           & Com2Sense (paired)       & general      & multiple-choice (2)  &    804 &        &    391 &    782 &    391 &    391 \\
SciQ          & SciQ                     & scientific   & multiple-choice (4)  &  11679 & \cmark &   1000 &   4000 &   1000 &   3000 \\
QuaRel        & QuaRel                   & qualitative  & multiple-choice (2)  &   1941 &        &    278 &    556 &    278 &    278 \\
QuaRTz        & QuaRTz                   & qualitative  & multiple-choice (2)  &   2696 &        &    384 &    768 &    384 &    384 \\
CycIC         & CycIC (mc)               & general      & multiple-choice (5)  &   6521 &        &    907 &   4535 &    907 &   3628 \\
ComVE         & ComVE (task A)           & general      & multiple-choice (2)  &  10000 &        &    997 &   1994 &    997 &    997 \\
CSQA2         & CommonsenseQA 2.0        & general      & boolean              &   9264 &        &   2541 &   2541 &   1225 &   1316 \\
SKD\_anno     & SKD (annotated)          &              & boolean              &   7980 &        &   1015 &   1015 &    803 &    212 \\
I2D2\_anno    & I2D2 (annotated)         &              & boolean              &  26206 &        &  13094 &  13094 &   6158 &   6936 \\
\textbf{Total} & & & & \textbf{194442} & & \textbf{29430} & \textbf{61329} & \textbf{20966} & \textbf{40363} \\
\midrule
\multicolumn{10}{c}{\textsc{Evaluation (unseen type 1)}} \\
\midrule
WSC           & WSC                      & general      & multiple-choice (2)  &      0 &        &    273 &    546 &    273 &    273 \\
COPA          & COPA                     & general      & multiple-choice (2)  &      0 &        &    500 &   1000 &    500 &    500 \\
NumerSense    & NumerSense               & quantitative & multiple-choice (11) &      0 &        &    200 &   2200 &    200 &   2000 \\
PROST         & PROST                    & physical     & multiple-choice (4)  &      0 &        &  18736 &  74944 &  18736 &  56208 \\
SpatialCS     & Spatial Commonsense      & physical     & boolean              &      0 &        &   1448 &   1448 &    724 &    724 \\
Rainier\_anno & Rainier (annotated)      &              & boolean              &      0 &        &    591 &    591 &    424 &    167 \\
\textbf{Total} & & & & \textbf{0} & & \textbf{21748} & \textbf{80729} & \textbf{20857} & \textbf{59872} \\

\midrule
\multicolumn{10}{c}{\textsc{Evaluation (unseen type 2)}} \\
\midrule
SWAG          & SWAG                     &              & multiple-choice (4)  &      0 &        &  20006 &  80024 &  20006 &  60018 \\
HellaSwag     & HellaSwag                &              & multiple-choice (4)  &      0 &        &  10042 &  40168 &  10042 &  30126 \\
CODAH         & CODAH                    &              & multiple-choice (4)  &      0 &        &   2776 &  11104 &   2776 &   8328 \\
SCT           & Story Cloze Test         &              & multiple-choice (2)  &      0 &        &   1871 &   3742 &   1871 &   1871 \\
$\alpha$NLI   & $\alpha$NLI              &              & multiple-choice (2)  &      0 &        &   1532 &   3064 &   1532 &   1532 \\
StrategyQA    & StrategyQA               &              & boolean              &      0 &        &    229 &    229 &    107 &    122 \\
CREAK         & CREAK                    &              & boolean              &      0 &        &   1371 &   1371 &    691 &    680 \\
\textbf{Total} & & & & \textbf{0} & & \textbf{37827} & \textbf{139702} & \textbf{37025} & \textbf{102677} \\
\bottomrule
\end{tabular}
}%
\caption{
    Datasets and statistics.
    Data sourced from commonsense KBs are listed under \textsc{Stage A training}, and data sourced from commonsense QA datasets are listed under \textsc{Stage B training}.
    The number in parentheses under the \textbf{Format} column represents the number of choices per question.
    The \textbf{Aug} column indicates whether LM-augmented falsehoods are generated for each dataset.
    The last three columns are the number of total, correct and incorrect statements in the development set.
    See \autoref{tab:datasets_stat} for more dataset statistics, and \autoref{tab:datasets_more} for full citations and sources for these datasets.
}
\label{tab:datasets}
\end{table*}

\begin{table*}[t]
\small
\setlength{\tabcolsep}{3pt}
\centering
\resizebox{0.6\textwidth}{!}{%
\begin{tabular}{l l rrrrrr}
\toprule
\textbf{Abbr.} & \textbf{Name} & \multicolumn{6}{c}{\textbf{Statement Length}} \\
& & min & median & 90\% & 95\% & 99\% & max \\
\midrule
\multicolumn{8}{c}{\textsc{Stage A training}} \\
\midrule
% The following content is automatically generated by `python dsstat.py`
Atomic2020    & Atomic2020               &    5 &   19 &   24 &   26 &   30 &   57 \\
GenericsKB    & GenericsKB               &    4 &   13 &   22 &   24 &   28 &   82 \\
\midrule
\multicolumn{8}{c}{\textsc{Stage B training (seen)}} \\
\midrule
OBQA          & OpenBookQA               &    5 &   16 &   29 &   36 &   56 &   74 \\
ARC\_e        & ARC (easy)               &    6 &   24 &   50 &   60 &   86 &  111 \\
ARC\_h        & ARC (hard)               &    7 &   30 &   59 &   70 &   94 &  138 \\
AI2Sci\_e     & AI2 Science (elem)       &    7 &   29 &   63 &   79 &  455 &  473 \\
AI2Sci\_m     & AI2 Science (middle)     &    7 &   24 &   58 &   72 &  511 &  536 \\
CSQA          & CommonsenseQA            &    5 &   18 &   28 &   32 &   43 &   73 \\
QASC          & QASC                     &    5 &   13 &   19 &   21 &   24 &   30 \\
PIQA          & Physical IQA             &    5 &   26 &   62 &   80 &  120 &  256 \\
SIQA          & Social IQA               &   10 &   28 &   38 &   41 &   51 &   70 \\
WG            & Winogrande               &   17 &   24 &   31 &   34 &   38 &   42 \\
C2S           & Com2Sense (paired)       &   12 &   24 &   34 &   38 &   44 &   55 \\
SciQ          & SciQ                     &    6 &   19 &   29 &   34 &   48 &   75 \\
QuaRel        & QuaRel                   &   15 &   39 &   62 &   80 &  101 &  107 \\
QuaRTz        & QuaRTz                   &    9 &   29 &   44 &   49 &   73 &   78 \\
CycIC         & CycIC (mc)               &    6 &   31 &   59 &   67 &   97 &  122 \\
ComVE         & ComVE (task A)           &    4 &   10 &   14 &   16 &   20 &   28 \\
CSQA2         & CommonsenseQA 2.0        &    5 &   14 &   24 &   29 &   38 &   58 \\
SKD\_anno     & SKD (annotated)          &   13 &   20 &   25 &   27 &   31 &   37 \\
I2D2\_anno    & I2D2 (annotated)         &    5 &   15 &   21 &   24 &   31 &   41 \\
\midrule
\multicolumn{8}{c}{\textsc{Evaluation (unseen type 1)}} \\
\midrule
WSC           & WSC                      &   10 &   22 &   33 &   39 &   45 &   48 \\
COPA          & COPA                     &   10 &   17 &   21 &   23 &   26 &   28 \\
NumerSense    & NumerSense               &    6 &   13 &   20 &   23 &   32 &   36 \\
PROST         & PROST                    &   17 &   42 &   63 &   68 &   78 &   78 \\
SpatialCS     & Spatial Commonsense      &    9 &   12 &   17 &   18 &   19 &   20 \\
Rainier\_anno & Rainier (annotated)      &    5 &   12 &   19 &   21 &   29 &   33 \\
\midrule
\multicolumn{8}{c}{\textsc{Evaluation (unseen type 2)}} \\
\midrule
SWAG          & SWAG                     &   12 &   30 &   46 &   52 &   67 &  148 \\
HellaSwag     & HellaSwag                &   15 &  103 &  134 &  140 &  149 &  181 \\
CODAH         & CODAH                    &    5 &   21 &   31 &   34 &   45 &   73 \\
SCT           & Story Cloze Test         &   29 &   56 &   69 &   72 &   78 &   89 \\
$\alpha$NLI   & $\alpha$NLI              &   17 &   35 &   44 &   47 &   53 &   65 \\
StrategyQA    & StrategyQA               &    6 &   14 &   20 &   22 &   25 &   30 \\
CREAK         & CREAK                    &    8 &   14 &   20 &   22 &   29 &   50 \\
\bottomrule
\end{tabular}
}%
\caption{
    More dataset statistics.
    This table shows the percentiles of statement lengths (as in number of T5 tokens) in each dataset.
}
\label{tab:datasets_stat}
\end{table*}
\begin{table*}[t]
\small
\setlength{\tabcolsep}{3pt}
\centering
\resizebox{\textwidth}{!}{%
\begin{tabular}{l l l p{220pt} c}
\toprule
\textbf{Abbr.} & \textbf{Name} & \textbf{Citation} & \textbf{Link} & \textbf{In Flan-T5?} \\
\midrule
\multicolumn{5}{c}{\textsc{Stage A training}} \\
\midrule
% The following content is automatically generated by `python dsstat.py`
Atomic2020    & Atomic2020               & \citet{Hwang2020COMETATOMIC2O}           & \url{https://allenai.org/data/atomic-2020}                                                        & yes \\
GenericsKB    & GenericsKB               & \citet{Bhakthavatsalam2020GenericsKBAK}  & \url{https://allenai.org/data/genericskb}                                                         & yes \\
\midrule
\multicolumn{5}{c}{\textsc{Stage B training (seen)}} \\
\midrule
OBQA          & OpenBookQA               & \citet{Mihaylov2018CanAS}                & \url{https://github.com/allenai/unifiedqa}                                                        & yes \\
ARC\_e        & ARC (easy)               & \citet{Clark2018ThinkYH}                 & \url{https://github.com/allenai/unifiedqa}                                                        & yes \\
ARC\_h        & ARC (hard)               & \citet{Clark2018ThinkYH}                 & \url{https://github.com/allenai/unifiedqa}                                                        & yes \\
AI2Sci\_e     & AI2 Science (elem)       & \citet{Clark2018ThinkYH}                 & \url{https://github.com/allenai/unifiedqa}                                                        & no  \\
AI2Sci\_m     & AI2 Science (middle)     & \citet{Clark2018ThinkYH}                 & \url{https://github.com/allenai/unifiedqa}                                                        & no  \\
CSQA          & CommonsenseQA            & \citet{Talmor2019CommonsenseQAAQ}        & \url{https://github.com/allenai/unifiedqa}                                                        & yes \\
QASC          & QASC                     & \citet{Khot2019QASCAD}                   & \url{https://github.com/allenai/unifiedqa}                                                        & yes \\
PIQA          & Physical IQA             & \citet{Bisk2019PIQARA}                   & \url{https://github.com/allenai/unifiedqa}                                                        & yes \\
SIQA          & Social IQA               & \citet{Sap2019SocialIC}                  & \url{https://github.com/allenai/unifiedqa}                                                        & yes \\
WG            & Winogrande               & \citet{Sakaguchi2019WINOGRANDEAA}        & \url{https://github.com/allenai/unifiedqa}                                                        & yes \\
C2S           & Com2Sense (paired)       & \citet{Singh2021COM2SENSEAC}             & \url{https://github.com/PlusLabNLP/Com2Sense/tree/master/data}                                    & yes \\
SciQ          & SciQ                     & \citet{Welbl2017CrowdsourcingMC}         & \url{https://allenai.org/data/sciq}                                                               & yes \\
QuaRel        & QuaRel                   & \citet{Tafjord2018QuaRelAD}              & \url{https://allenai.org/data/quarel}                                                             & yes \\
QuaRTz        & QuaRTz                   & \citet{Tafjord2019QuaRTzAO}              & \url{https://allenai.org/data/quartz}                                                             & yes \\
CycIC         & CycIC (mc)               & --                                      & \url{https://leaderboard.allenai.org/cycic/submissions/get-started}                               & no  \\
ComVE         & ComVE (task A)           & \citet{Wang2020SemEval2020T4}            & \url{https://github.com/wangcunxiang/SemEval2020-Task4-Commonsense-Validation-and-Explanation}    & no  \\
CSQA2         & CommonsenseQA 2.0        & \citet{Talmor2021CommonsenseQA2E}        & \url{https://github.com/allenai/csqa2/tree/master/dataset}                                        & no  \\
SKD\_anno     & SKD (annotated)          & \citet{West2021SymbolicKD}               & \url{https://github.com/peterwestai2/symbolic-knowledge-distillation/tree/main/purification_code} & no  \\
I2D2\_anno    & I2D2 (annotated)         & \citet{Bhagavatula2022I2D2IK}            & \url{https://gengen.apps.allenai.org}                                                             & no  \\
\midrule
\multicolumn{5}{c}{\textsc{Evaluation (unseen type 1)}} \\
\midrule
WSC           & WSC                      & \citet{Levesque2011TheWS}                & \url{https://huggingface.co/datasets/winograd_wsc}                                                & yes \\
COPA          & COPA                     & \citet{Gordon2011SemEval2012T7}          & \url{https://huggingface.co/datasets/super_glue}                                                  & yes \\
NumerSense    & NumerSense               & \citet{Lin2020BirdsHF}                   & \url{https://github.com/INK-USC/NumerSense/tree/main/data}                                        & yes \\
PROST         & PROST                    & \citet{ArocaOuellette2021PROSTPR}        & \url{https://huggingface.co/datasets/corypaik/prost}                                              & yes \\
SpatialCS     & Spatial Commonsense      & \citet{Liu2022ThingsNW}                  & \url{https://github.com/xxxiaol/spatial-commonsense}                                              & no  \\
Rainier\_anno & Rainier (annotated)      & \citet{Liu2022RainierRK}                 & \url{https://github.com/liujch1998/rainier}                                                       & no  \\
\midrule
\multicolumn{5}{c}{\textsc{Evaluation (unseen type 2)}} \\
\midrule
SWAG          & SWAG                     & \citet{Zellers2018SWAGAL}                & \url{https://github.com/rowanz/swagaf/tree/master/data}                                           & yes \\
HellaSwag     & HellaSwag                & \citet{Zellers2019HellaSwagCA}           & \url{https://github.com/rowanz/hellaswag/tree/master/data}                                        & yes \\
CODAH         & CODAH                    & \citet{chen2019codah}                    & \url{https://github.com/Websail-NU/CODAH/tree/master/data}                                        & yes \\
SCT           & Story Cloze Test         & \citet{Mostafazadeh2016ACA}              & \url{https://cs.rochester.edu/nlp/rocstories/}                                                    & yes \\
$\alpha$NLI   & $\alpha$NLI              & \citet{Bhagavatula2019AbductiveCR}       & \url{https://leaderboard.allenai.org/anli/submissions/get-started}                                & yes \\
StrategyQA    & StrategyQA               & \citet{Geva2021DidAU}                    & \url{https://github.com/eladsegal/strategyqa/tree/main/data/strategyqa}                           & yes \\
CREAK         & CREAK                    & \citet{Onoe2021CREAKAD}                  & \url{https://github.com/yasumasaonoe/creak/tree/main/data/creak}                                  & yes \\
\bottomrule
\end{tabular}
}%
\caption{
    More dataset details.
    We show the link from which we retrieved each dataset, and whether each dataset is included in the training data of Flan-T5.
}
\label{tab:datasets_more}
\end{table*}

\section{More Details on Method}
\label{sec:method_more}

\subsection{Training Objectives}

\paragraph{Binary classification loss.}
We defined the binary classification loss as
\begin{align*}
& \loss_{\text{bin}}(x_i, y_i) = \\
&\qquad -y_i \log s(x_i) - (1 - y_i) \log (1 - s(x_i)).
\end{align*}
To account for the fact that there are usually more incorrect statements than correct ones in the data produced from multiple-choice datasets, we divide this loss by the number of statements with the same correctness label in the same statement group.
Therefore, the binary classification loss for the whole batch is
\begin{align*}
& \loss_{\text{bin}} = \\
&\quad \frac{1}{B_G} \sum_{j=1}^{B_G} \sum_{y \in \{0, 1\}} \frac{\sum_{c=1}^{C_j}{ \mathbb{I}[y_{jc} = y] \loss_{\text{bin}}(x_{jc}, y_{jc}) }}{\sum_{c=1}^{C_j}{ \mathbb{I}[y_{jc} = y] }},
\end{align*}
where $C_j$ is the number of statements in statement group $X_j$, $x_{jc}$ is the $c$th statement in $X_j$, and $\mathbb{I}$ is the indicator function.
% \wenya{Added some notation explanations here. A small issue: should we start from $c=1$ instead of $c=0$?} \gary{Fixed.}

\paragraph{Multi-class loss.}
We defined the multi-class loss as
\begin{align*}
\loss_{\text{mc}}(X_j) &= -\log \frac{\exp z(x_{j*})}{\sum_{c=1}^{C_j} \exp z(x_{jc})}.
\end{align*}
The multi-class loss for the whole batch is
\begin{align*}
\loss_{\text{mc}} &= \frac{1}{B_G} \sum_{j=1}^{B_G} \loss_{\text{mc}}(X_j).
\end{align*}

\paragraph{Supervised contrastive loss.}
We defined the supervised contrastive loss as
\begin{align*}
\loss_{\text{ctr}}(x_i, y_i) &= -\log \frac{\sum_{k \in \mathcal{P}(i)} e^{\frac{\text{cos}(\mathbf{h}(x_i), \mathbf{h}(x_k))}{\tau}}}{\sum_{k \in \mathcal{P}(i) \cup \mathcal{N}(i)} e^{\frac{\text{cos}(\mathbf{h}(x_i), \mathbf{h}(x_k)) }{\tau}}}.
\end{align*}
The supervised contrastive loss for the whole batch is
\begin{align*}
\loss_{\text{ctr}} &= \frac{1}{B_S} \sum_{i=1}^{B_S} \loss_{\text{ctr}}(x_i, y_i).
\end{align*}

% \subsection{Model architecture}

\subsection{Calibration}
\label{sec:calibration_more}

Our calibration is a post-hoc strategy and does not affect the task performance metrics we report in \S\ref{sec:results}.
This is because applying our calibration method -- temperature scaling -- does not affect the relative order of plausibility scores assigned to a given set of statements:
\begin{itemize}
\item For tasks with multiple-choice questions (\S\ref{sec:results_probsolving}), calibration does not affect the argmax prediction for the above reason.
\item For commonsense knowledge filtering (\S\ref{sec:results_knowledge}), calibration does not affect the TPR/FPR numbers at each corresponding decision point, again for the above reason, so the ROC curves are valid.
\item For True/False judgment problems (\S\ref{sec:results_probsolving} and \S\ref{sec:results_chatgpt}), calibration does not move the plausibility scores across the decision boundary. We use logit $z=0.0$ (or equivalently, plausibility score $s=0.5$) as the True/False boundary. A positive (or negative) logit remains positive (or negative) after applying the temperature.
\end{itemize}

\section{More Details on Experimental Setup}
\label{sec:app-exp-setup}

\autoref{tab:hypers} shows the hyperparamter settings for training \methodname{}.
These values are obtained from some moderate hyperparameter tuning, and we did not do extensive search due to training cost.

For tokenization, the T5 tokenizer tokenizes input so that it ends with the EOS token \texttt{</s>} (token ID = 1). We manually configured the LLaMA tokenizer so that its output ends with the EOS token \texttt{</s>} (token ID = 2), and does not contain the BOS token \texttt{<s>} (token ID = 1). Models are trained for $S = 50k$ steps with $B_G = 64$ statement groups per batch, using the Adam optimizer \citep{Kingma2014AdamAM} with learning rate $\eta = 1 \times 10^{-5}$ for T5 encoder and $\eta = 2 \times 10^{-6}$ for LLaMA.
We train models with the Huggingface Transformers and Accelerate libraries \citep{Wolf2019HuggingFacesTS, accelerate}. For memory efficiency, during training, each statement is truncated to 128 tokens (which can accommodate more than 99\% of the statements; see \autoref{tab:datasets_stat}) and each statement group is capped to four statements.

\begin{table*}[t]
\small
\centering
\begin{tabular}{c c l}
\toprule
\textbf{Symbol} & \textbf{Value} & \textbf{Description} \\
\midrule
$L$ & 128 & Max number of tokens in the input statement \\
$B_G$ & 64 & Number of statement groups per batch \\
$C$ & 4 & Max number of statements in each group, during training \\
$B_S$ & 256 & Max number of statements per batch, during training \\
$S$ & 50,000 & Total number of training steps in each stage \\
$\eta_{\text{T5}}$ & $1 \times 10^{-5}$ & Learning rate for \methodname{} with \model{T5} encoder backbone \\
$\eta_{\text{LLaMA}}$ & $2 \times 10^{-6}$ & Learning rate for \methodname{} with \model{LLaMA} backbone\\
$\alpha$ & 1.0 & Weight of binary classification loss \\
$\beta$ & 1.0 & Weight of multiple-choice loss \\
$\gamma$ & 0.1 & Weight of supervised contrastive loss \\
$\tau$ & 0.05 & Temperature in supervised contrastive loss \\
\bottomrule
\end{tabular}
\caption{
    Hyperparameter settings.
    % \nascomment{this feels like appendix material (but do point to it in the main text)}
}
\label{tab:hypers}
\end{table*}

\subsection{Definition of Metrics}
\label{sec:def_metrics}

% \nascomment{cut from here ...}
\paragraph{Multiple-choice accuracy.}
For multiple-choice benchmarks, we report the multiple-choice accuracy:
\begin{align*}
Acc_{\text{mc}} &= \frac{1}{|\Ds|} \sum_{X_j \in \Ds}{\mathbb{I}[x_{j*} = \argmax_{x_{jc} \in X_j}{s(x_{jc})}]}.
\end{align*}

\paragraph{Boolean accuracy.}
The boolean accuracy is defined as
\begin{align*}
Acc_{\text{bool}} &= \frac{1}{|\Ds|} \sum_{(x_i, y_i) \in \Ds}{\mathbb{I}\big[y_i = \mathbb{I}[z(x_i) > 0]\big]}.
\end{align*}

% \hanna{agree with noah; cut the formulas - they are clear; if you like to keep them, move them to appendix. keep the he following few sentences about the datasets you use boolean accuracy.} 
Boolean accuracy is applicable to balanced boolean benchmarks where there are roughly equal true and false statements (e.g., CommonsenseQA 2.0, Spatial Commonsense, StrategyQA, CREAK).
Generally it is not a good metric for multiple-choice benchmarks and unbalanced boolean benchmarks.

\paragraph{AUROC and AP.}
For unbalanced boolean benchmarks (e.g., LM-generated knowledge filtering datasets), accuracy may not faithfully capture the model's performance.
Instead, the metrics we use are the area under the ROC curve (AUROC) and the average precision (AP) for selecting the True statements.
Statements are ranked based on their assigned raw scores, so that different score thresholds can be selected to construct the ROC and Precision-Recall curves.
Aside from unbalanced boolean benchmarks, AUROC and AP are also applicable to multiple-choice and balanced boolean benchmarks.

\paragraph{Calibration.}
To measure how well the verifier-predicted score reflects its confidence, we measure the ECE \citep{Naeini2015ObtainingWC} on the boolean benchmarks.
% \nascomment{... to here}
ECE is computed as
\begin{align}
\text{ECE} &= \sum_{m=1}^{M}{\frac{|B_m|}{|\Ds|} \cdot \Big| \text{Acc}(B_m) - \text{Score}(B_m) \Big|} \nonumber \\
&= \sum_{m=1}^{M} \frac{|B_m|}{|\Ds|} \cdot \Big| \frac{1}{|B_m|} \sum_{(x_i, y_i) \in B_m} \mathbb{I}[y_i = 1] \nonumber \\
&\qquad\qquad - \frac{1}{|B_m|} \sum_{(x_i, y_i) \in B_m} s(x_i) \Big|,
\label{eqn:ece}
\end{align}
% \wenya{Explanations for notations are missing here.}
% \gary{Added some notation explanations.}
where $M$ is the number of bins which bucket data points with similar predictions, and $B_m \subseteq \Ds$ is the subset of data points that fall into the $m$-th bin.
We use $M=10$ equal-sized bins when computing ECE.

\subsection{Details on Baseline Models}
\label{sec:baselines_more}

\paragraph{SKD Critic.}
\citet{West2021SymbolicKD} trained a critic model that filters incorrect commonsense knowledge generated by their symbolic knowledge distillation (SKD) method.
This critic model is based on \ckpt{RoBERTa-large} \citep{Liu2019RoBERTaAR} and is finetuned on $8k$ \model{GPT-3}-generated commonsense knowledge sentences with human-annotated true/false labels.
The model predicts a $[0, 1]$ score $s$ which we use as the final score, and we let the logit $z = \sigma^{-1}(s)$.

\paragraph{I2D2 Critic.}
\citet{Bhagavatula2022I2D2IK} trained a critic model that filters incorrect commonsense knowledge generated by their \model{I2D2} method.
This critic model is based on \ckpt{RoBERTa-large} \citep{Liu2019RoBERTaAR} and is finetuned on $12k$ \model{I2D2}-generated commonsense knowledge sentences with human-annotated true/false labels.
Given an input statement, the model predicts two logits:  $t$ for the True label and $f$ for the False label.
We let the logit $z = t - f$ and the score $s = \sigma(t - f)$.
We use the critic model trained in the final iteration (i.e., ``Iter 2'' in I2D2\footnote{\url{https://gengen.apps.allenai.org/}}).

\paragraph{UnifiedQA-v2.}
\model{UnifiedQA-v2} \citep{Khashabi2022UnifiedQAv2SG} is a general-purpose QA model trained on datasets with a variety of input formats, including boolean datasets.
When the input is a declarative statement, the model is trained to output either ``yes'' or ``no''.
We use this feature of the model and make it act as a commonsense statement verifier.
For an input statement, we compute the logits received by ``yes'' and ``no'' in the decoder, denoted as $t$ and $f$, respectively.
We let the logit $z = t - f$ and the score $s = \sigma(t - f)$.
We use the largest version of this model, \ckpt{UnifiedQA-v2-11b}.\footnote{\url{https://huggingface.co/allenai/unifiedqa-v2-t5-11b-1251000}}

\paragraph{Entailer.}
\model{Entailer} \citep{Tafjord2022EntailerAQ} is a model trained to construct proof trees for scientific commonsense hypotheses.
This multi-angle model can be used in three ways: (1) given a hypothesis, generate a set of premises that may entail it; (2) given a hypothesis, predict a score that reflects the model's belief in it; (3) given a hypothesis and set of premises, predict a score that reflects whether there is a valid entailment between them.
We use (2) as a commonsense statement verifier.
The model predicts a $[0, 1]$ score $s$ which we use as the final score, and we let the logit $z = \sigma^{-1}(s)$.
We use the largest version of this model, \ckpt{Entailer-11b}.\footnote{\url{https://huggingface.co/allenai/entailer-11b}}

\paragraph{GPT-3.5.}
\model{GPT-3.5} \citep{gpt35docs} is a series of general-purpose autoregressive decoder-only LMs.
To make it act as a commonsense verifier, we use the following input prompt:
\begin{align*}
& \texttt{\scriptsize Question: Based on commonsense knowledge, is the following} \\ 
& \texttt{\scriptsize statement correct? Please answer yes or no.} \\
& \texttt{\scriptsize Statement: \{statement\}} \\
& \texttt{\scriptsize Answer:}
\end{align*}
We query the OpenAI Completions API\footnote{\url{https://platform.openai.com/docs/api-reference/completions}} with this prompt and compute the logits received by `` Yes'' and `` No'' in the next-token prediction, denoted as $t$ and $f$, respectively.
We let the logit $z = t - f$ and the score $s = \sigma(t - f)$.
We experimented with several prompt formats and found the one presented above to have the best performance, and in most cases, `` Yes'' and `` No'' together receive most of the probability mass during next-token prediction.
We also experimented with several models in the \model{GPT-3} \citep{Brown2020LanguageMA} and \model{GPT-3.5} series, and found \ckpt{GPT-3.5 (text-davinci-002)} to work the best.
% We did not benchmark \model{ChatGPT} \citep{chatgptblog} and \model{GPT-4} \citep{gpt4}, because their APIs do not provide token logits. \yejin{Presumably we can still use ChatGPT/GPT-4 in binary classification mode for the purpose of computing accuracy or knowledge filtering, to not sure if this reasoning is tight enough… i wonder if we can give a different or additional reason?} \gary{Added ChatGPT baseline, see below.}

\noindent
Additionally, we report a baseline where the (negated) language modeling perplexity is used for commonsense plausibility.
Note that the plausibility scores derived this way are not normalized, and we only use them for ranking purposes.
For this baseline, we use \ckpt{GPT-3.5 (text-davinci-002)} as the base model, and name it as ``PPL (\model{GPT-3.5})''.

\paragraph{ChatGPT and GPT-4.}
\model{ChatGPT} \citep{chatgptblog} and \model{GPT-4} \citep{gpt4} are optimized for chat.
To make them act as a commonsense verifier, we use the same input prompt as for \model{GPT-3.5}, without the \texttt{``Answer:''} line.
We query the OpenAI Chat API\footnote{\url{https://platform.openai.com/docs/api-reference/chat}} with this prompt in a user message, and obtain the first token of the assistant message in the response.
Besides this zero-shot setting, we additionally report a few-shot chain-of-thought \citep{Wei2022ChainOT} setting with 5 in-domain examples, formatted as additional user-assistant message pairs prior to the query user message.

\noindent
Since the API does not provide token logits, we let the score $s = 1.0$ when this token is ``Yes'', and $s = 0.0$ when this token is ``No''.
In the unlikely case that this token is neither, we let $s = 0.5$.
We add a small random noise to the score.
This is to arbitrate potentially multiple positive predictions within statement groups from multiple-choice QA problems, and to enable plotting the ROC and precision-recall curves.
Note that this is not an ideal solution and may cause under-estimation of \model{ChatGPT} and \model{GPT-4}'s performance.

\paragraph{Flan-T5.}
\model{Flan-T5} \citep{Chung2022ScalingIL} is a series of sequence-to-sequence LMs instruction-finetuned on massive number of tasks.
To make it act as a commonsense verifier, we use the same input prompt as for GPT-3.5.
%\begin{align*}
%\scriptsize
%& \texttt{\scriptsize Question: Based on commonsense knowledge, is the following statement correct? Please answer yes or no.} \\
%& \texttt{\scriptsize Statement: \{statement\}} \\
%& \texttt{\scriptsize Answer:}
%\end{align*}
We compute the logits received by ``yes'' and ``no'' in the first token prediction in the decoder, denoted as $t$ and $f$, respectively.
We let the logit $z = t - f$ and the score $s = \sigma(t - f)$.
We experimented with several prompt formats and found the one presented above to have the best performance, and in most cases, ``yes'' and ``no'' together receive most of the probability mass during the token prediction.
We use the largest version of this model, \ckpt{Flan-T5-XXL}.\footnote{\url{https://huggingface.co/google/flan-t5-xxl}}
Note that some unseen benchmarks are in the training data of \model{Flan-T5}; see \autoref{tab:datasets_more} for details on data contamination.

\section{More Evaluation Results}
\label{sec:eval_more}

\begin{figure*}[t]
\centering
% This image is generated by `python visualize_acc.py`
\includegraphics[width=0.69\linewidth]{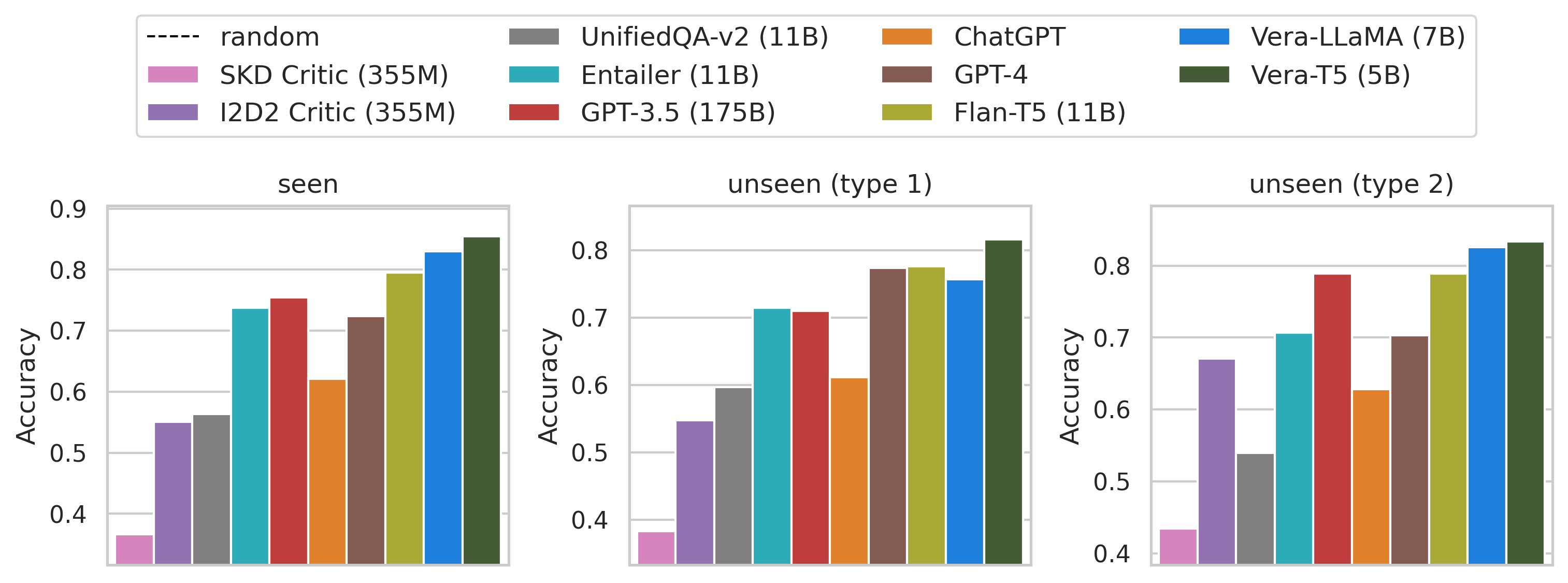}
% This image is generated by `python visualize_boolean.py full`
\includegraphics[width=0.69\linewidth]{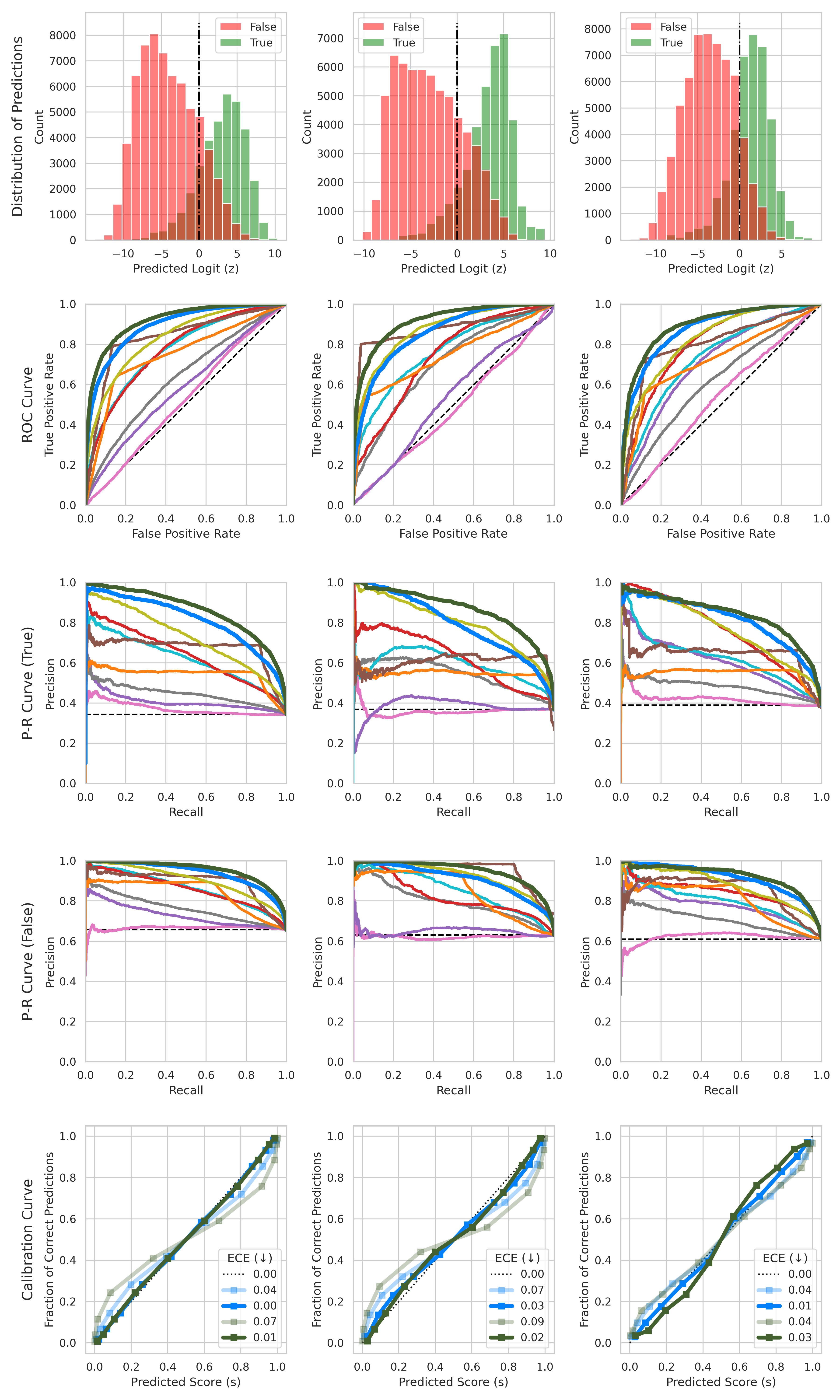}
\caption{
    Results on problem-solving with \methodname{} on seen and unseen benchmarks.
    Average results on the development sets are reported.
    Accuracy across different parts (seen, unseen (type 1), unseen (type 2)) are not directly comparable due to different underlying benchmarks.
    For calibration curves, curves with saturated colors are results after applying post hoc calibration (\S\ref{sec:calibration}), while curves with faded colors are results from the raw logits.
}
\label{fig:results_probsolving_full}
% \vspace{-16pt}
\end{figure*}

\begin{figure*}[t]
\centering
% This image is generated by `python visualize_knowledge.py full`
\includegraphics[width=0.88\linewidth]{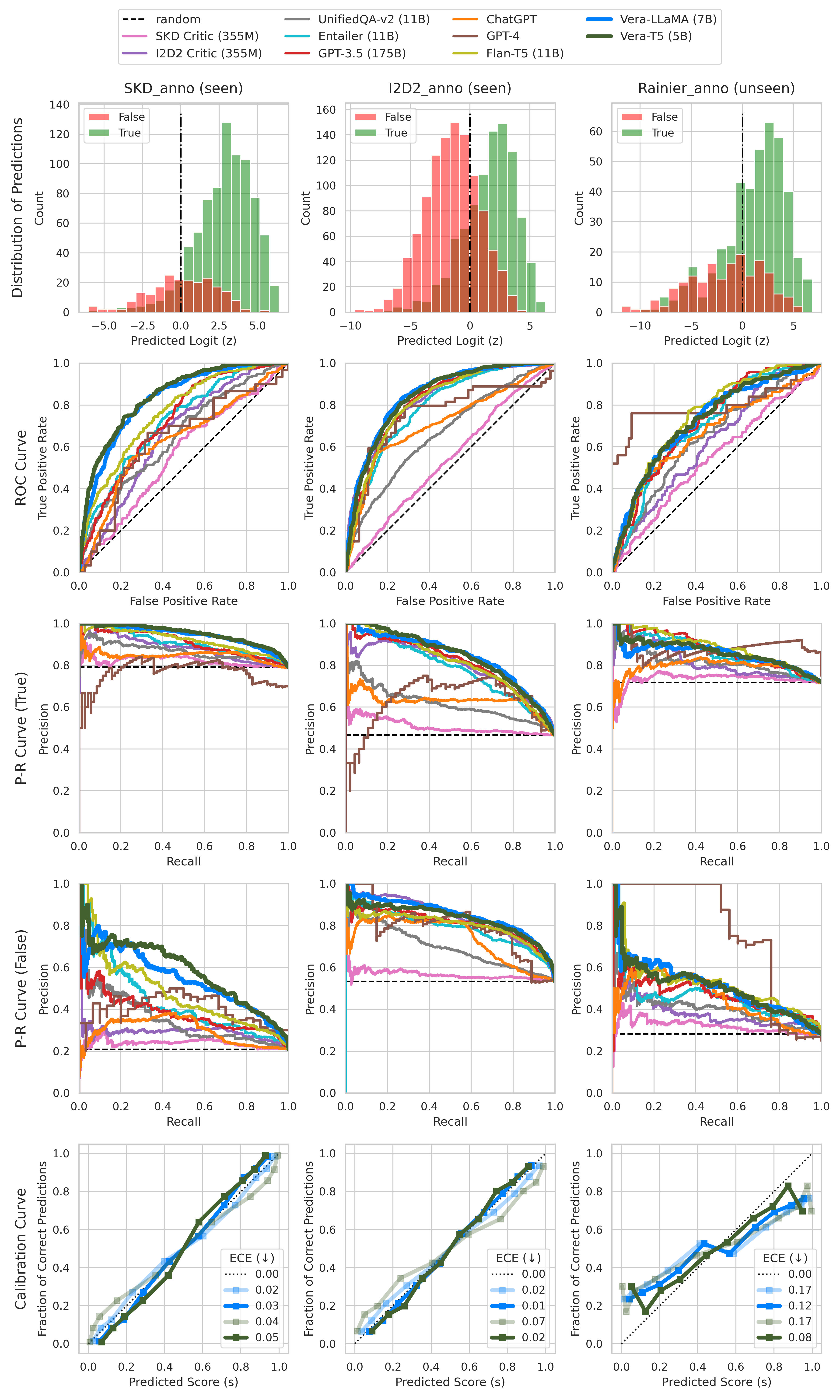}
\caption{
    Results for filtering LM-generated commonsense knowledge with \methodname{}.
    Results on the development sets are reported.
}
\label{fig:results_knowledge_full}
% \vspace{-16pt}
\end{figure*}

\autoref{fig:results_probsolving_full} is an expansion of \autoref{fig:results_probsolving} and additionally shows the precision-recall curves on problem-solving benchmarks.
\autoref{tab:results_seen}, \autoref{tab:results_unseen}, and \autoref{tab:results_ood} show the per-dataset breakdown of the accuracy numbers in \autoref{fig:results_probsolving_full}.
\autoref{fig:results_knowledge_full} is an expansion of \autoref{fig:results_knowledge} and additionally shows the precision-recall curves on knowledge-filtering benchmarks.
\autoref{tab:results_gpi_more} shows the per-dataset breakdown of the accuracy numbers in \autoref{tab:results_gpi}.

\begin{table*}[t]
\setlength{\tabcolsep}{2pt}
\centering
\resizebox{\textwidth}{!}{%
% This table is automatically generated by `python visualize_acc.py`
\begin{tabular}{l cccccccccc cccccccccc} 
\toprule
\textbf{Dataset} $\rightarrow$ & \textbf{All} & \textbf{MC} & \textbf{Bool} & OBQA & ARC\_e & ARC\_h & AI2Sci\_e & AI2Sci\_m & CSQA & QASC & PIQA & SIQA & WG & C2S & SciQ & QuaRel & QuaRTz & CycIC
 & ComVE & CSQA2 \\
\midrule
SKD Critic (355M) & 36.64 & 35.96 & 47.60 & 27.60 & 29.12 & 23.08 & 25.20 & 28.00 & 20.15 & 12.42 & 53.86 & 39.20 & 50.28 & 51.41 & 27.30 & 53.60 & 55.73 & 25.80 & 52.56 & 47.60 \\
I2D2 Critic (355M) & 55.03 & 55.75 & 43.65 & 44.80 & 55.61 & 35.79 & 55.28 & 50.40 & 61.51 & 45.25 & 67.36 & 56.45 & 55.56 & 63.17 & 55.10 & 61.51 & 60.16 & 35.72 & 88.26 & 43.65 \\
UnifiedQA-v2 (11B) & 56.33 & 56.34 & 56.05 & 54.60 & 48.77 & 39.46 & 48.78 & 43.20 & 44.23 & 32.61 & 63.98 & 52.10 & 70.64 & 75.96 & 42.20 & 81.65 & 71.35 & 48.29 & 83.65 & 56.05 \\
Entailer (11B) & 73.79 & 74.90 & 56.00 & 74.40 & 81.93 & 64.88 & 77.24 & 82.40 & 67.81 & 57.56 & 78.78 & 64.33 & 77.11 & 82.86 & 76.90 & 85.97 & 76.56 & 52.81 & 96.89 & 56.00 \\
PPL (\model{GPT-3.5}) & 66.02 & 66.02 & -- & 45.20 & 70.12 & 45.15 & 69.92 & 62.70 & 61.92 & 57.02 & 81.55 & 51.23 & 71.37 & 72.56 & 86.80 & 71.22 & 70.91 & 53.31 & 85.37 & -- \\
GPT-3.5 (175B) & 75.41 & 76.34 & 60.55 & 74.20 & 85.79 & 68.90 & 84.55 & 80.80 & 66.91 & 62.85 & 84.17 & 65.30 & 72.53 & 81.33 & 86.00 & 83.09 & 76.04 & 51.60 & 97.39 & 60.55 \\
ChatGPT & 62.11 & 61.52 & 71.65 & 60.80 & 65.44 & 57.19 & 63.41 & 68.00 & 39.64 & 42.01 & 67.36 & 52.20 & 61.33 & 76.73 & 60.70 & 74.10 & 72.66 & 29.66 & 93.08 & 71.65 \\
\quad + 5-shot CoT & 65.19 & 65.19 & -- & 62.40 & 77.33 & 62.88 & 72.36 & 69.84 & 46.52 & 47.52 & 68.59 & 52.25 & 59.98 & 82.14 & 69.77 & 65.83 & 72.14 & 42.56 & 90.98 & -- \\
GPT-4 & 72.35 & 71.81 & 81.00 & 76.00 & 69.00 & 72.00 & 80.00 & 80.00 & 43.00 & 44.00 & 73.00 & 57.00 & 77.00 & 94.00 & 70.00 & 86.00 & 80.00 & 53.00 & 95.00 & 81.00 \\
\quad + 5-shot CoT & 74.96 & 74.96 & -- & 79.80 & 79.00 & 70.00 & 87.00 & 89.11 & 31.31 & 44.00 & 80.00 & 67.00 & 82.18 & 91.92 & 80.00 & 82.00 & 87.00 & 57.00 & 92.08 & -- \\
Flan-T5 (11B) & 79.50 & 80.58 & 62.25 & 79.60 & 85.79 & 71.24 & 86.99 & 81.60 & 69.21 & 64.58 & 83.95 & 73.23 & 84.69 & 84.40 & 80.80 & 92.81 & 82.03 & 69.90 & 98.40 & 62.25 \\
\midrule
\methodname{}-LLaMA (7B) & 82.99 & 84.18 & 63.85 & 80.20 & 84.39 & 75.92 & 88.62 & 82.40 & 76.17 & 71.38 & 85.91 & 79.89 & 87.92 & 83.63 & 90.00 & 92.09 & 80.99 & 89.42 & 97.99 & 63.85 \\
\methodname{}-T5 (5B) & 85.51 & 86.57 & 68.60 & 83.20 & 88.07 & 78.60 & 93.50 & 86.40 & 77.97 & 73.33 & 88.47 & 80.14 & 92.42 & 85.93 & 88.80 & 93.88 & 84.90 & 91.73 & 97.79 & 68.60 \\
\bottomrule
\end{tabular}
}%
% \vspace{-4pt}
\caption{
    Results on \textbf{seen} benchmarks.
    Accuracy on the development set is reported.
}
\label{tab:results_seen}
\vspace{12pt}
\resizebox{0.7 \textwidth}{!}{%
% This table is automatically generated by `python visualize_acc.py`
\begin{tabular}{l cccccccc}
\toprule
\textbf{Dataset} $\rightarrow$ & \textbf{All} & \textbf{MC} & \textbf{Bool} & WSC & COPA & NumerSense & PROST & SpatialCS \\
\midrule
SKD Critic (355M) & 38.34 & 35.83 & 48.41 & 54.21 & 53.00 & 11.50 & 24.60 & 48.41 \\
I2D2 Critic (355M) & 54.79 & 54.43 & 56.22 & 80.59 & 72.80 & 35.00 & 29.35 & 56.22 \\
UnifiedQA-v2 (11B) & 59.73 & 55.10 & 78.25 & 71.79 & 81.20 & 35.00 & 32.40 & 78.25 \\
Entailer (11B) & 71.47 & 68.05 & 85.15 & 86.08 & 92.40 & 51.00 & 42.70 & 85.15 \\
GPT-3.5 (175B) & 71.03 & 70.73 & 72.24 & 85.71 & 87.00 & 66.50 & 43.70 & 72.24 \\
ChatGPT & 61.20 & 54.69 & 87.22 & 73.26 & 58.80 & 47.50 & 39.20 & 87.22 \\
GPT-4 & 77.40 & 71.75 & 100.00 & 85.00 & 64.00 & 69.00 & 69.00 & 100.00 \\
Flan-T5 (11B) & 77.62 & 73.22 & 95.23 & 90.48 & 93.00 & 57.50 & 51.90 & 95.23 \\
\midrule
\methodname{}-LLaMA (7B) & 75.71 & 74.06 & 82.32 & 94.14 & 91.80 & 65.00 & 45.30 & 82.32 \\
\methodname{}-T5 (5B) & 81.65 & 78.70 & 93.44 & 94.51 & 93.40 & 66.50 & 60.40 & 93.44 \\
\bottomrule
\end{tabular}
}%
% \vspace{-4pt}
\caption{
    Results on \textbf{unseen (type 1)} benchmarks.
    Accuracy on the development set is reported.
}
\label{tab:results_unseen}
\vspace{12pt}
\resizebox{0.8 \textwidth}{!}{%
% This table is automatically generated by `python visualize_acc.py`
\begin{tabular}{l cccccccccc}
\toprule
\textbf{Dataset} $\rightarrow$ & \textbf{All} & \textbf{MC} & \textbf{Bool} & SWAG & HellaSwag & CODAH & SCT & $\alpha$NLI & StrategyQA & CREAK \\
\midrule
SKD Critic (355M) & 43.40 & 40.11 & 51.62 & 26.95 & 30.45 & 29.35 & 62.75 & 51.04 & 50.66 & 52.59 \\
I2D2 Critic (355M) & 67.11 & 70.42 & 58.84 & 72.15 & 53.30 & 67.30 & 88.24 & 71.08 & 52.40 & 65.28 \\
UnifiedQA-v2 (11B) & 53.95 & 52.83 & 56.77 & 31.75 & 36.60 & 49.00 & 82.04 & 64.75 & 49.34 & 64.19 \\
Entailer (11B) & 70.72 & 70.63 & 70.94 & 52.45 & 47.65 & 80.70 & 94.39 & 77.94 & 60.26 & 81.62 \\
GPT-3.5 (175B) & 78.87 & 80.21 & 75.53 & 73.40 & 70.40 & 85.05 & 95.56 & 76.63 & 62.88 & 88.18 \\
ChatGPT & 62.83 & 56.21 & 79.39 & 43.70 & 42.95 & 56.75 & 77.34 & 60.31 & 67.69 & 91.10 \\
GPT-4 & 70.29 & 66.20 & 80.50 & 57.00 & 40.00 & 66.00 & 93.00 & 75.00 & 70.00 & 91.00 \\
Flan-T5 (11B) & 78.89 & 80.52 & 74.81 & 69.20 & 64.55 & 89.60 & 98.45 & 80.81 & 61.14 & 88.48 \\
\midrule
\methodname{}-LLaMA (7B) & 82.56 & 86.11 & 73.71 & 79.30 & 83.90 & 88.95 & 98.61 & 79.77 & 62.45 & 84.97 \\
\methodname{}-T5 (5B) & 83.37 & 86.66 & 75.13 & 76.30 & 85.90 & 88.60 & 98.56 & 83.94 & 65.07 & 85.19 \\
\bottomrule
\end{tabular}
}%
% \vspace{-4pt}
\caption{
    Results on \textbf{unseen (type 2)} benchmarks.
    Accuracy on the development set is reported.
}
\label{tab:results_ood}
\end{table*}
\begin{table*}[t]
\setlength{\tabcolsep}{2pt}
\centering
\resizebox{\textwidth}{!}{%
\begin{tabular}{lll cccccccccc c}
\toprule
Generator & Filter & QA & \textbf{Avg} & OBQA & ARC\_e & ARC\_h & AI2Sci\_e & AI2Sci\_m & CSQA & QASC & PIQA & SIQA & WG \\
\midrule
-- & -- & UnifiedQA-large & 60.45 & 70.20 & 69.12 & 55.85 & 69.11 & 64.80 & 61.43 & 43.09 & 63.66 & 53.84 & 53.35 \\
GPT-3 (davinci) & -- & UnifiedQA-large & 67.44 & 74.60 & 75.44 & 64.55 & 69.92 & 72.80 & 70.19 & 63.82 & 67.74 & 58.70 & 56.59 \\
\midrule
GPT-3 (davinci) & \methodname{} & UnifiedQA-large & \textbf{70.67} & \textbf{77.60} & \textbf{80.00} & \textbf{67.56} & \textbf{78.05} & \textbf{78.40} & \textbf{71.91} & \textbf{66.20} & \textbf{70.35} & \textbf{59.37} & \textbf{57.22} \\
\bottomrule
\end{tabular}
}%
\vspace{16pt}
\resizebox{\textwidth}{!}{%
\begin{tabular}{lll cccccccccc c}
\toprule
Generator & Filter & QA & \textbf{Avg} & OBQA & ARC\_e & ARC\_h & AI2Sci\_e & AI2Sci\_m & CSQA & QASC & PIQA & SIQA & WG \\
\midrule
-- & -- & UnifiedQA-large & 60.45 & 70.20 & 69.12 & 55.85 & 69.11 & 64.80 & 61.43 & 43.09 & 63.66 & 53.84 & 53.35 \\
Rainier-large & -- & UnifiedQA-large & 61.78 & 69.40 & 66.84 & 52.84 & 68.29 & 57.60 & \textbf{68.30} & 54.86 & 65.51 & 56.81 & 57.38 \\
\midrule
Rainier-large & \methodname{} & UnifiedQA-large & \textbf{64.88} & \textbf{73.40} & \textbf{71.05} & \textbf{57.19} & \textbf{73.98} & \textbf{67.20} & \textbf{68.30} & \textbf{55.51} & \textbf{67.52} & \textbf{56.96} & \textbf{57.70} \\
\bottomrule
\end{tabular}
}%
% \vspace{-4pt}
\caption{
    Results of introducing \methodname{} into the Generated Knowledge Prompting pipeline \citep{Liu2021GeneratedKP}.
    Accuracy on the development set is reported.
}
\label{tab:results_gpi_more}
\end{table*}

\section{Further Analysis}
\label{sec:further_analysis}

\begin{figure*}[t]
\centering
\includegraphics[width=0.8\linewidth]{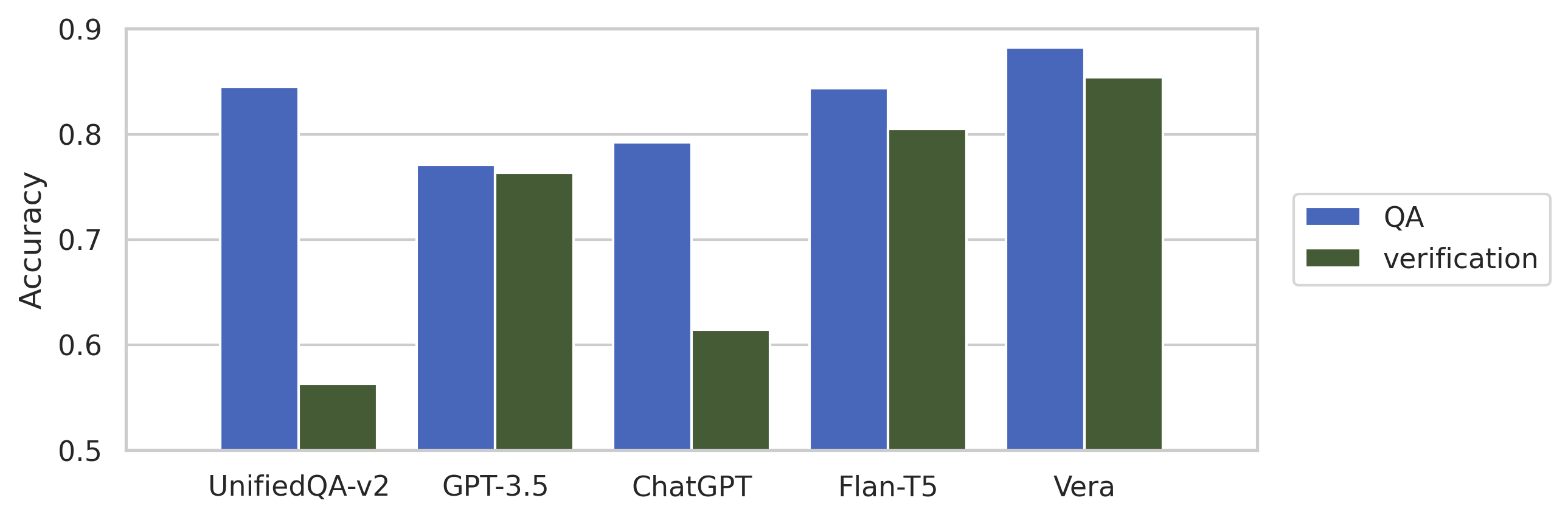}
\caption{
    Comparing verification and QA, the two different formats for problem-solving tasks.
    Average accuracy on the development sets of the seen multiple-choice benchmarks is reported.
    We use \ckpt{text-davinci-002} as \model{GPT-3.5} here, and \ckpt{gpt-3.5-turbo-0301} as \model{ChatGPT}.
    % We do not have data for ChatGPT solving problems in verification format, because its API does not return token logits or scores.
    \methodname{} in QA format actually means a \model{T5} model finetuned on the same seen multiple-choice data as \methodname{}.
}
\label{fig:results_format}
% \vspace{-16pt}
\end{figure*}

\paragraph{Format: Verification vs. QA.}
% \label{sec:format}
% \nascomment{I don't understand why this question is interesting?  this seems to be a question about doing well on QA accuracy, which is not the point of the paper.  I think you need to start off by saying something like, ``On the surface, the verification format we use in this work appears to be an alternative to formats typically used in question answering.  It is natural to ask whether it is suitable for that purpose (QA).''  then delete the next two sentences.}
% \hanna{I don't think this section and analyses add much to your story. Specially, using verification format vs. qa format for other models. I propose to move this to appendix; then, in the accuracy report section, have a footnote or a few sentences that mentions if you train VERA with the same data for accuracy format you would gain blah, which is higher.  then mention vera is capable of doing verification, which has these advantages.}
In this paper, we have been using the verification format to approach problem-solving tasks. 
But do we lose something when compared to using the QA format?
In \autoref{fig:results_format} we compare how well existing models can solve problems in the verification format and the QA format.
Verification format does fall behind QA format, especially with models trained exclusively in QA format (i.e., \model{UnifiedQA-v2}).
We also trained a sequence-to-sequence model in QA format on the same multiple-choice data as \methodname{}.
It leads \methodname{} by 1.5\% on seen multiple-choice benchmarks.
We hypothesize that this is because verification models only see one option at a time, whereas QA models can see all choices of a problem at the same time and thus can do comparative ranking.

\noindent
In addition to the performance loss, a verification model does lose the generative capability possessed by some QA models that are generative (e.g., UnifiedQA in \citet{Khashabi2020UnifiedQACF}), and it has to run $C$ times to solve a $C$-way multiple-choice problem, whereas QA models (e.g., UnifiedQA in \citet{Khashabi2020UnifiedQACF}, Unicorn in \citet{Lourie2021UNICORNOR}) need to run only once.

\noindent
However, verification models can perform some tasks that generative QA models cannot cover.
They can classify the correctness of declarative statements, without having to convert them into questions in the first place.
They can also reflect on the answer produced by a generative QA model, and provide a level of confidence.
We argue that verification models and generative QA models have different best-application scenarios and are sometimes complementary to each other.
% \wenya{Shall we emphasize that our method outperforms either format?}
% \gary{The above are all zero-shot baselines (except UQA-v2). I think a fair comparison would be to train a QA model on the same datasets, and compare it with \methodname{}. The QA model might still perform better, but let's see.}

\end{document}